\documentclass{article} % For LaTeX2e
\usepackage{iclr2026_conference_modified_forArxiv,times}

\usepackage{amsmath,amsfonts,bm}

\def\eqref#1{equation~\ref{#1}}
\def\1{\bm{1}}

\DeclareMathAlphabet{\mathsfit}{\encodingdefault}{\sfdefault}{m}{sl}
\SetMathAlphabet{\mathsfit}{bold}{\encodingdefault}{\sfdefault}{bx}{n}

\usepackage{hyperref}
\usepackage{url}
\usepackage{booktabs}       % professional-quality tables
\usepackage{amsfonts}       % blackboard math symbols
\usepackage{nicefrac}       % compact symbols for 1/2, etc.
\usepackage{microtype}      % microtypography
\usepackage{xcolor}         % colors
\usepackage{amsmath}

\usepackage{graphicx}
\usepackage{subcaption}
\usepackage{multirow}
\usepackage{changepage,threeparttable}
\usepackage{wrapfig}
\usepackage{comment}

\title{Boosting Open Set Recognition Performance through Modulated Representation Learning}

\author{Amit Kumar Kundu\textsuperscript{1}, Vaishnavi S Patil\textsuperscript{2} \& Joseph Jaja\textsuperscript{1}
\\
\textsuperscript{1}Department of Electrical and Computer Engineering\\
\textsuperscript{2}Department of Computer Science\\
University of Maryland\\
\texttt{\{amit314,vspatil,josephj\}@umd.edu} \\
}

\iclrfinalcopy % Uncomment for camera-ready version, but NOT for submission.
\begin{document}

\maketitle

\begin{abstract}
The open set recognition (OSR) problem aims to identify test samples from novel semantic classes that are not part of the training classes, a task that is crucial in many practical scenarios. However, the existing OSR methods use a constant scaling factor (the temperature) to the logits before applying a loss function, which hinders the model from exploring both ends of the spectrum in representation learning -- from instance-level to semantic-level features. In this paper, we address this problem by enabling temperature-modulated representation learning using a set of proposed temperature schedules, including our novel negative cosine schedule. Our temperature schedules allow the model to form a coarse decision boundary at the beginning of training by focusing on fewer neighbors, and gradually prioritizes more neighbors to smooth out the rough edges. This gradual task switching leads to a richer and more generalizable representation space. While other OSR methods benefit by including regularization or auxiliary negative samples, such as with mix-up, thereby adding a significant computational overhead, our schedules can be folded into any existing OSR loss function with no overhead. We implement the novel schedule on top of a number of baselines, using cross-entropy, contrastive and the ARPL loss functions and find that it boosts both the OSR and the closed set performance in most cases, especially on the tougher semantic shift benchmarks. Project codes will be available.
%\href{https://anonymous.4open.science/r/NegCosSch-4516/}{\textit{here}}.
%\footnote{The project codes are available at \href{https://anonymous.4open.science/r/NegCosSch-4516/}{https://anonymous.4open.science/r/NegCosSch-4516/}.}.
\end{abstract}

\section{Introduction}\label{sec:intro}
Deep learning models have shown impressive performance by learning useful representations particularly for tasks involving the classification of examples into categories present in the training dataset, also known as the closed set. 
However during inference, in many practical scenarios, test samples may appear from unknown classes (termed as the open set), which were not a part of the training set. 
Hence, a more realistic task known as the open set recognition (OSR) (\citet{scheirer2012toward,chen2020learning}) aims to simultaneously flag the test samples from unknown classes while accurately classifying examples from the known classes, requiring strong generalization beyond the support of training data.
%We will interchangeably refer the closed set samples as known samples and the open set samples as the unknown ones.
%to the examples from known classes as in-distribution (InD) examples and examples from unknown classes as the out-of-distribution (OoD) examples.

\begin{figure}[t]
  \centering
  \begin{subfigure}[b]{0.245\textwidth} 
    \centering
    \includegraphics[width=\textwidth]{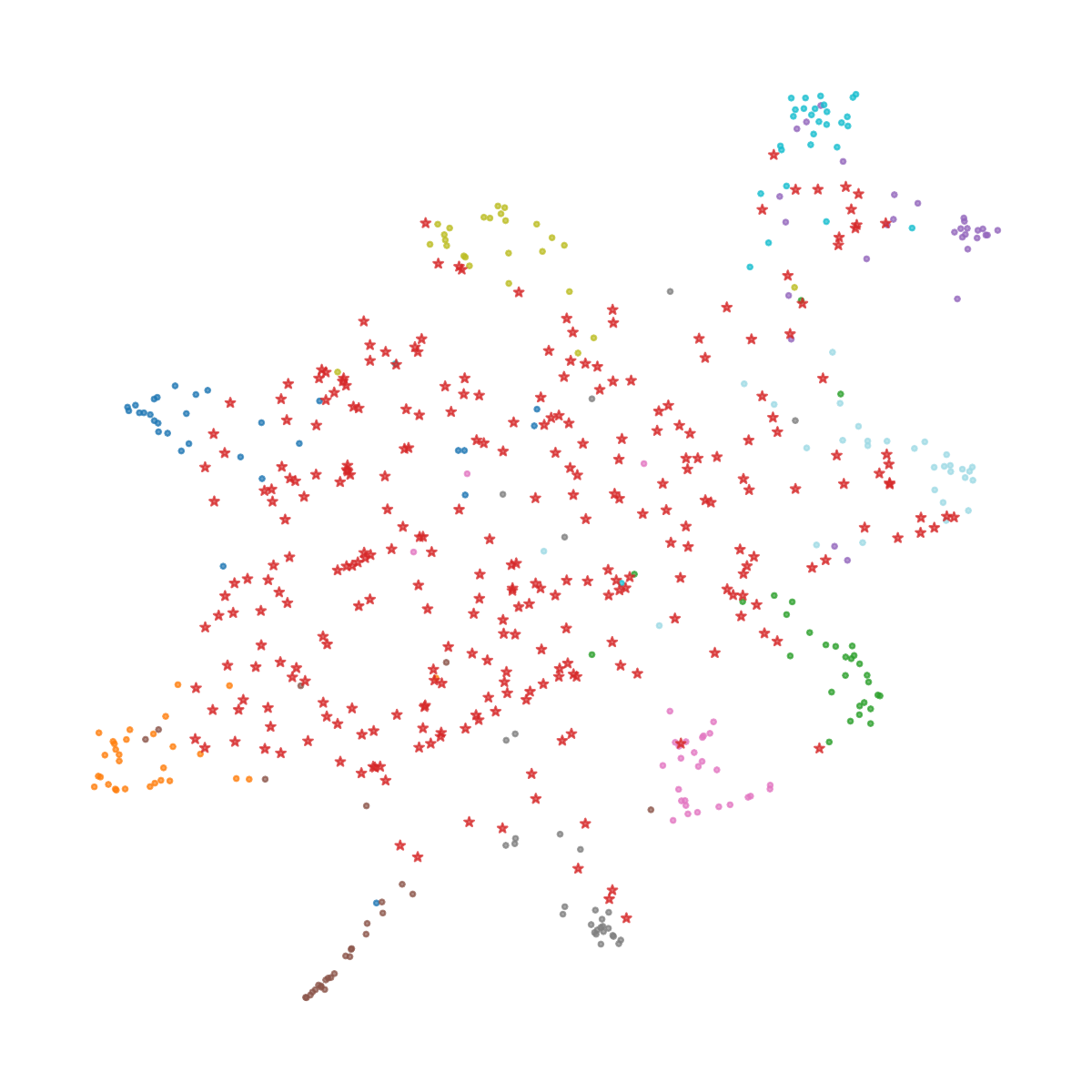}
    \caption{$\tau=0.5$} % 
    \label{fig0:subfig_a} 
  \end{subfigure}
  \begin{subfigure}[b]{0.245\textwidth} 
    \centering
    \includegraphics[width=\textwidth]{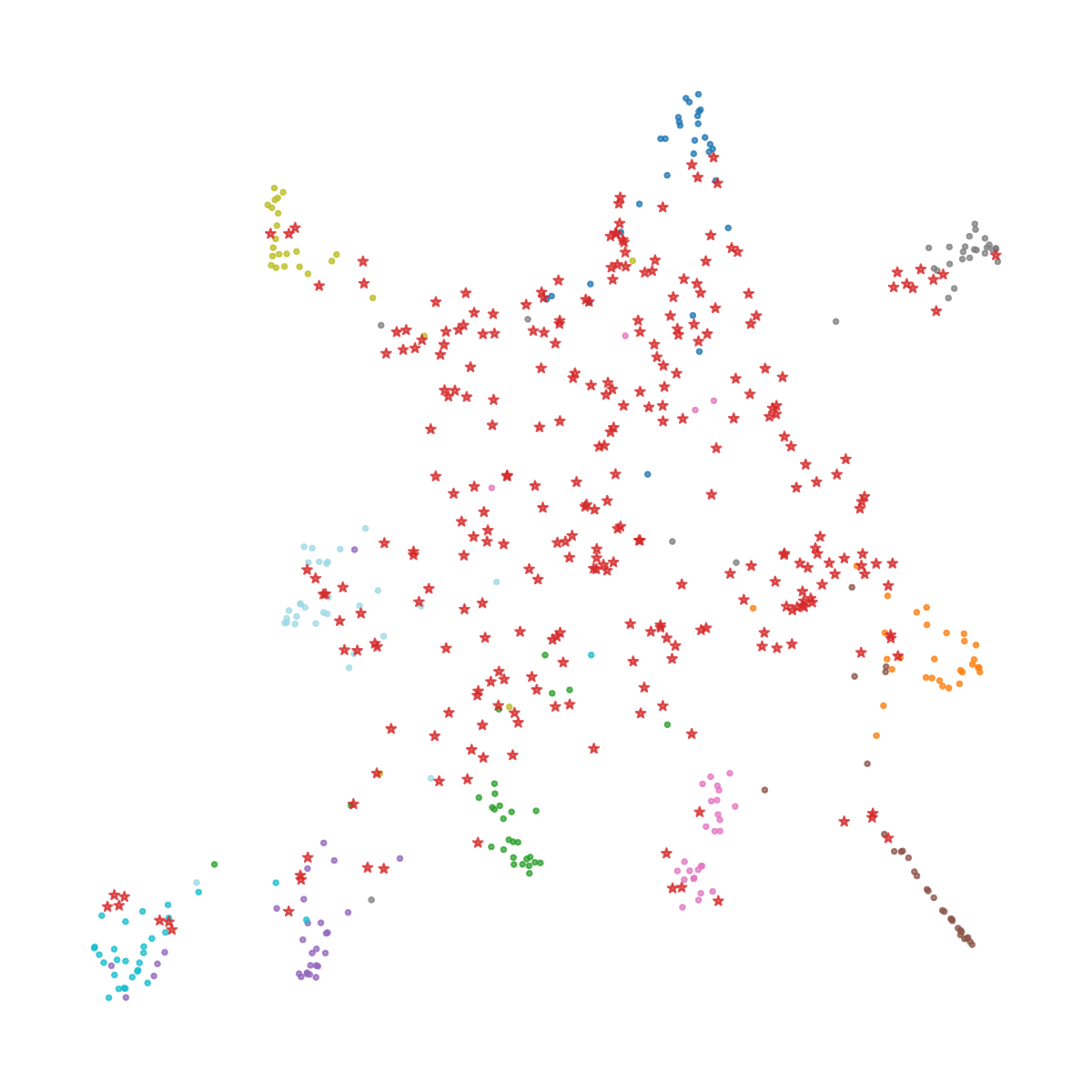}
    \caption{$\tau=1.0$} % 
    \label{fig0:subfig_b} 
  \end{subfigure}
  \begin{subfigure}[b]{0.245\textwidth} 
    \centering
    \includegraphics[width=\textwidth]{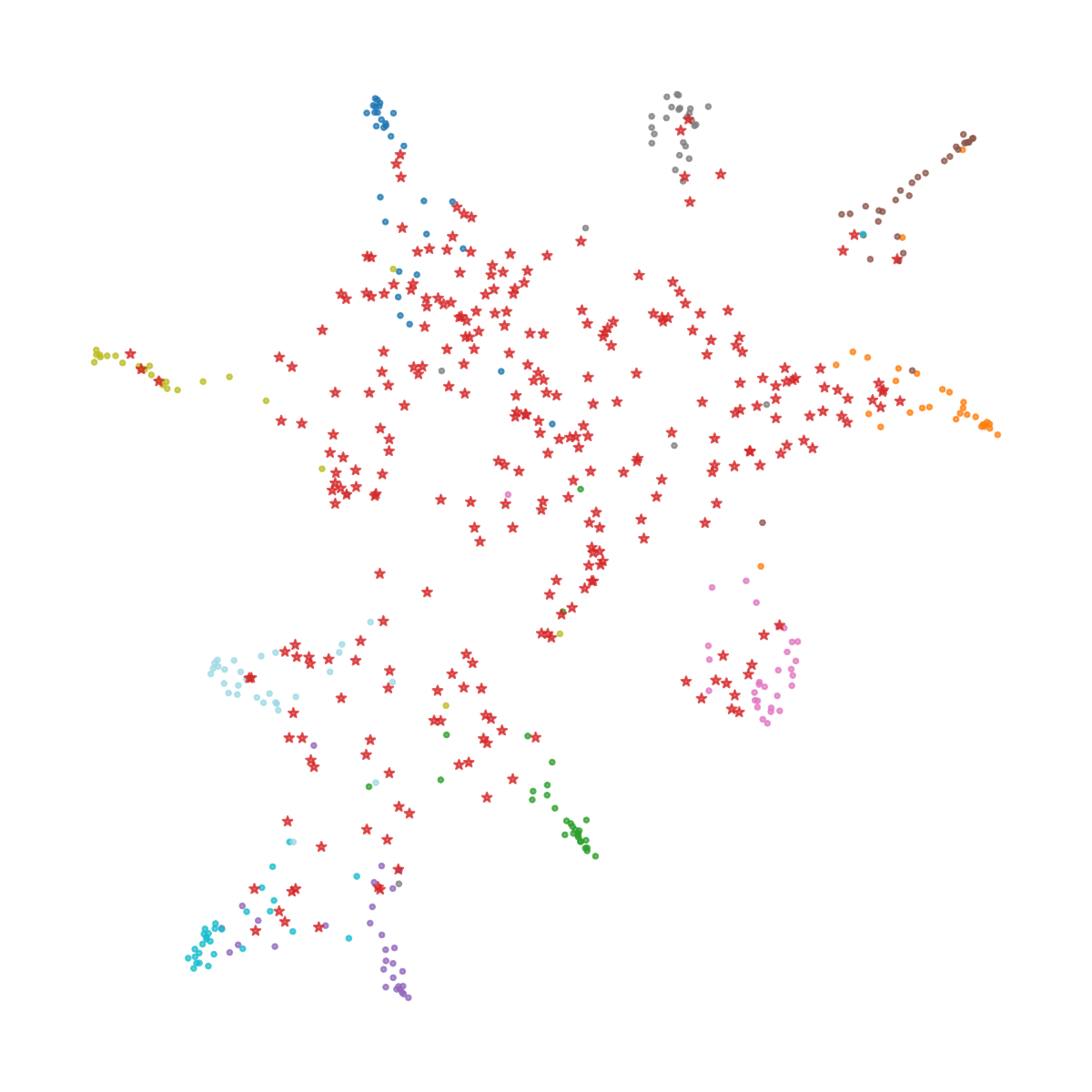}
    \caption{$\tau=2.0$} % 
    \label{fig0:subfig_c} 
  \end{subfigure}
  \begin{subfigure}[b]{0.245\textwidth} 
    \centering
    \includegraphics[width=\textwidth]{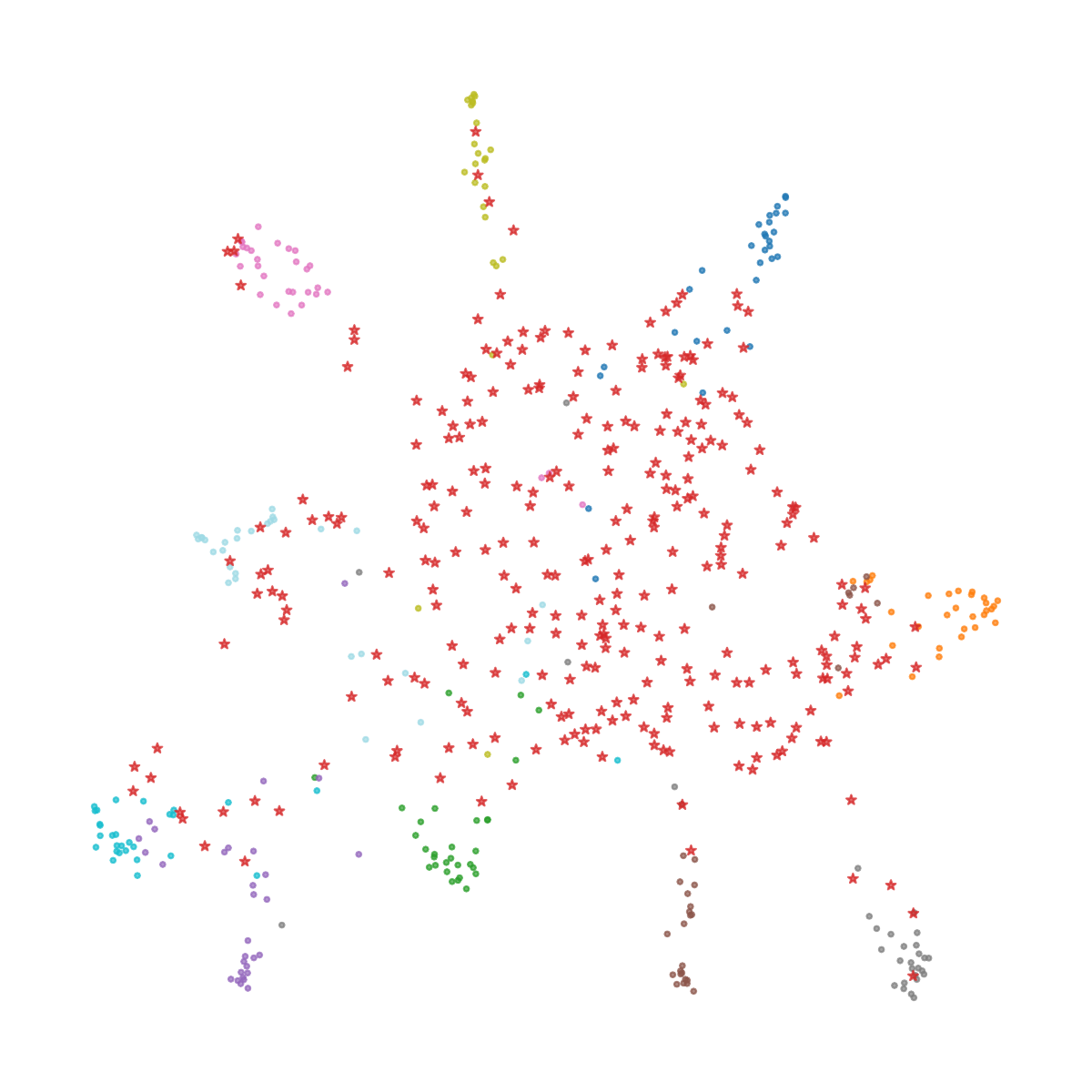}
    \caption{CosSch} % 
    \label{fig0:subfig_d} 
  \end{subfigure}\\
  \begin{subfigure}[b]{0.275\textwidth} 
    \centering
    \includegraphics[width=\textwidth]{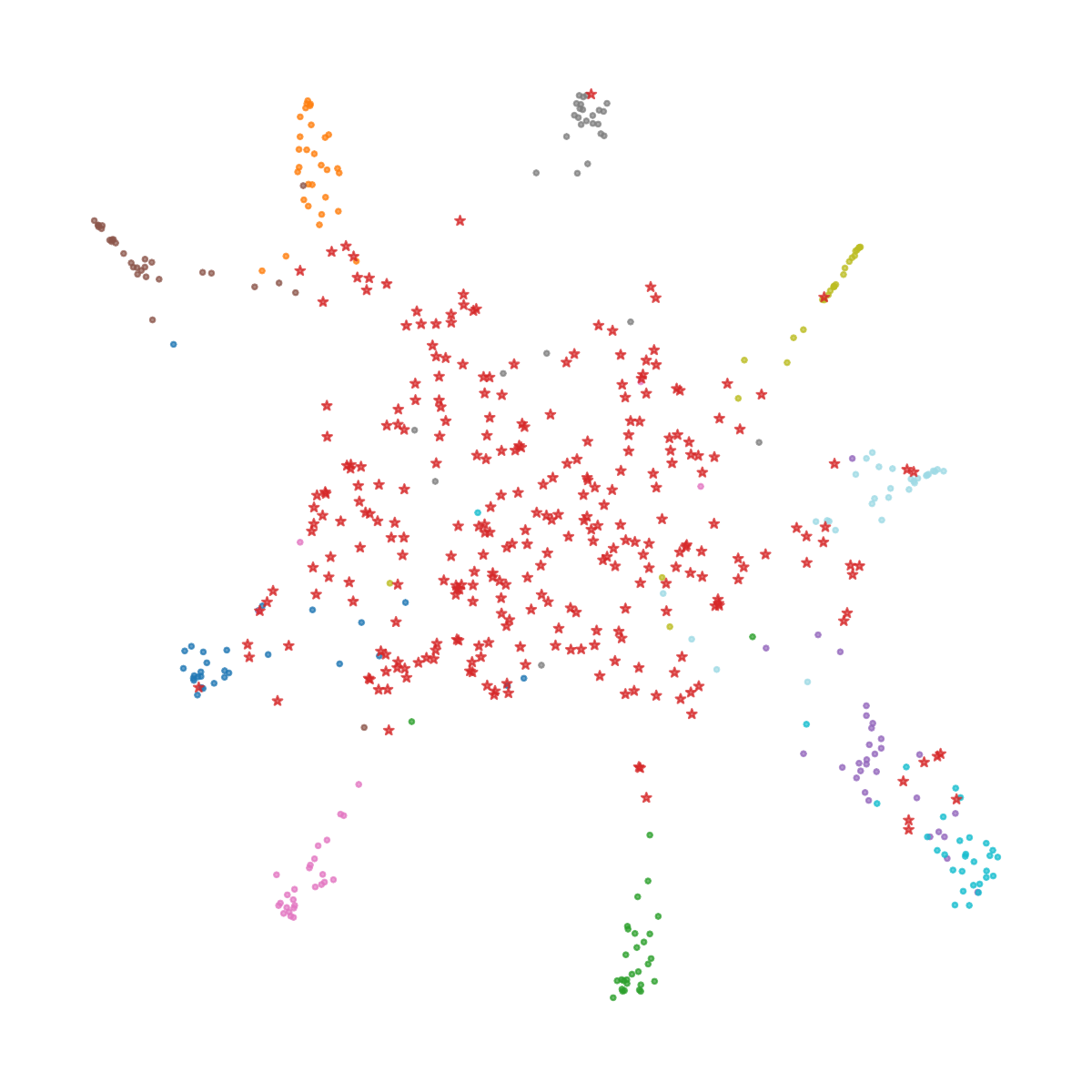}
    \caption{NegCosSch -- starting of the final period} % 
    \label{fig0:subfig_e} 
  \end{subfigure}~~~
  \begin{subfigure}[b]{0.275\textwidth} 
    \centering
    \includegraphics[width=\textwidth]{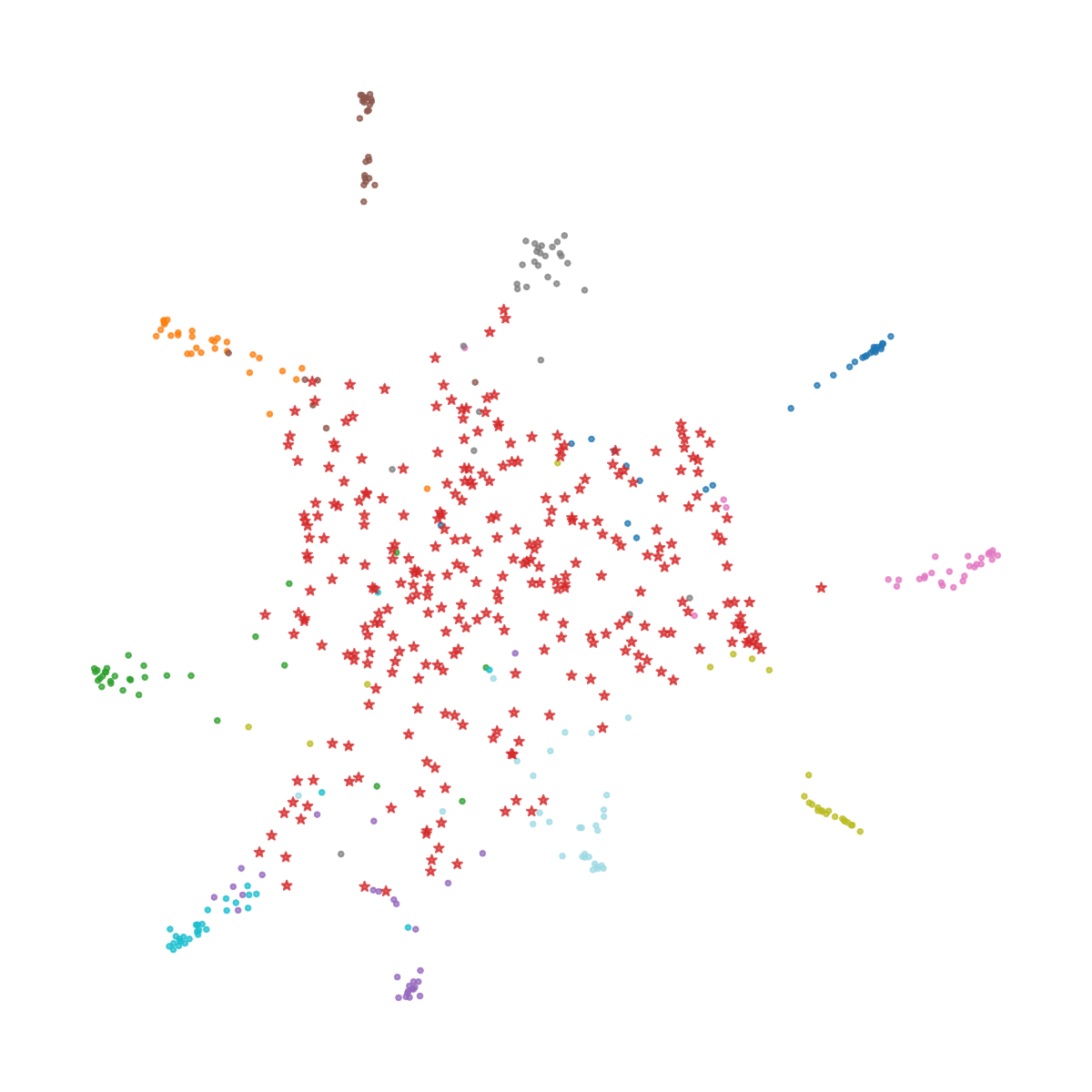}
    \caption{NegCosSch -- as training progresses} % 
    \label{fig0:subfig_f} 
  \end{subfigure}~~~
  \begin{subfigure}[b]{0.275\textwidth} 
    \centering
    \includegraphics[width=1.37\textwidth]{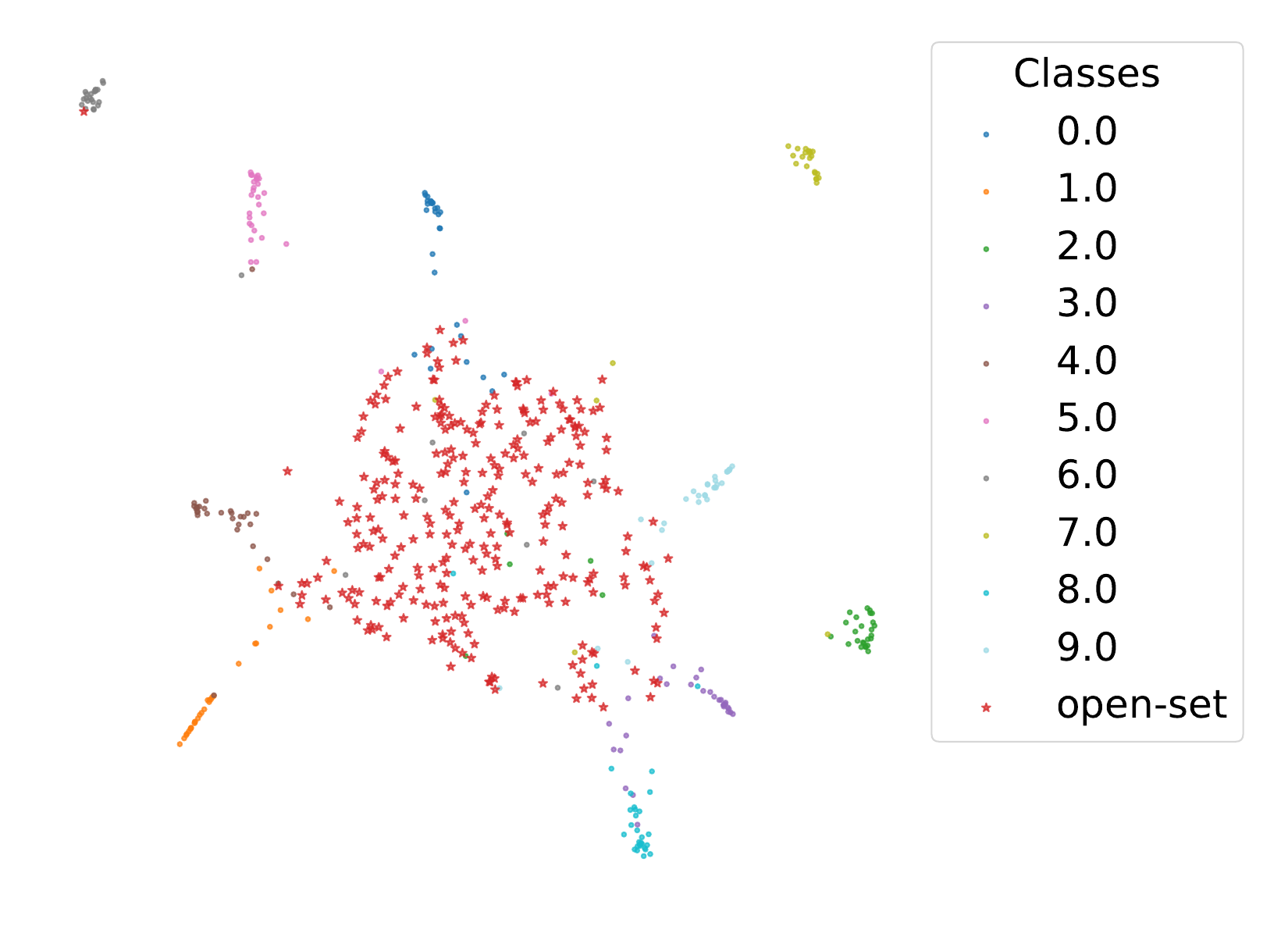}
    \caption{NegCosSch -- when training finishes} % 
    \label{fig0:subfig_g} \hfill
  \end{subfigure}
  \caption{UMAP projection of representation spaces for different temperature schedules on 10 classes of the Caltech-UCSD-Birds dataset. (a)- (c) show representations for constant temperatures ($\tau$). For lower $\tau$, the representations of unknown classes are distributed and so are the representations of known classes, leading to a sharp decision boundary. For higher $\tau$, the representations of known classes are  more compact, making the decision boundary smoother. However, unknown samples overlap with the clusters of known classes. Mid value of $\tau$ achieves a trade-off but does not gain the benefits of both ends. (e)- (g) show representations of our temperature schedule NegCosSch as the training progresses. A lower $\tau$ at the start leads to a coarse decision and the model gradually makes the classes more compact and the unknown representations are pushed away. Finally, (d) show the representation space for a previous schedule CosSch, which is better than fixed temperatures but not as compact as our NegCosSch. Experimental details appear in the Appendix.}\label{fig:0}
\end{figure}

Most of the early research attempts either model the unknown classes as long-tailed distributions \citep{vignotto2018extreme, bendale2016towards}, generate synthetic samples using generative models \citep{ge2017generative, neal2018open, chen2021adversarial, moon2022difficulty} or with mix-up \citep{chen2021adversarial,xu2023contrastive, li2024all, zhou2021learning} to represent novel classes, or train a secondary model with a separate objective, such as VAEs that include reconstruction based objective (\cite{oza2019c2ae, yoshihashi2019classification, zhou2024contrastive}). The synthetic examples may not generalize well to a variety of unknown classes, whereas training generative or secondary models or with mix-up are computationally demanding and often require higher memory. Later methods add regularization \citep{zhou2021learning, chen2021adversarial, chen2020learning} to explicitly bound the open space risks. In essence, these methods create more empty regions in the representation space hoping that unknown representations lie in those regions. Forcing the creation of empty spaces does not result in an improved OSR as the tougher unknown samples lie close to known classes in representation space, incurring significant similarity between them and reducing the effectiveness of such methods.

\citet{vaze2022openset} establish new OSR baselines by training models with optimal design choices and argue that a well-trained closed set classifier achieves an improved OSR performance, where the unknown samples exhibit lower max-logit scores. This essentially has motivated the next generation of OSR methods to learn even better representations for improving performance through a better loss function, such as the contrastive loss (\citet{khosla2020supervised,chen2020simple}) with sample mix-up (\citet{xu2023contrastive, verma2018manifold,zhang2017mixup}) and by adding different regularization schemes (\citet{zhou2024contrastive,bahavan2025sphor,li2025unlocking, wang2025backmix}). 

Moreover, the regular OSR benchmarks commonly used are small in scale. 
In this regard, semantic shift benchmarks (SSBs) are proposed by \citet{vaze2022openset} on fine-grained datasets, having more classes with varying levels of OSR difficulty.
Therefore, the methods that demonstrate improvement on smaller datasets but involve either data generation, mix-up, or training secondary models are unsuitable for the larger benchmarks as training a well-performing base model on them requires a significant compute and memory. Most of the latest research does not use these benchmarks. 
This necessitates the development of an advanced representation learning scheme that impose minimal computational overhead.

To achieve this, we need to explore the inner mechanisms of the losses that are the basis of most OSR methods, such as the cross-entropy (CE) and the contrastive loss. These loss functions compute probabilities by applying a temperature scaling to the logits- the model's raw outputs- where the temperature coefficient adjusts the sharpness of resulting probability distributions. It is the key parameter to control the learned features for both losses. Prior works \citep{wang2021understanding, zhang2022dual, kukleva2023temperature, zhang2021temperature} demonstrate that a lower temperature encourages instance-specific representations while a higher value encourages semantic ones.
However, a fixed temperature throughout the training prevents the model from exploring both ends of this learning spectrum. In this regard, \citet{kukleva2023temperature} study the benefits of learning both instance-level and semantic features primarily in the closed-set scenarios using self-supervised contrastive learning for long-tailed datasets using a cosine temperature schedule (TS), but the impact of temperature scaling or a TS remains largely unexamined for novel classes.

In OSR, learning a representation space that provides both instance-specific and semantic features is also crucial to achieve improved open set and closed set performance. 
In this research, we analyze the representation space for different temperature scaling factors on both losses in an open set scenario. Based on the analysis, we propose novel temperature schedules for temperature modulated representation learning. 
We find temperature modulation with the proposed schedules is beneficial to create more compact clusters for representing the closed set classes, while keeping open set examples more distant from these clusters, resulting in overall improved representations.

%The recent OSR methods shift gears to contrastive training due to their ability in learning better representations for many tasks . The contrastive loss computes scaled similarity between pair-wise samples, with the scaling factor called the temperature. Research works in \citet{zhang2022dual, kukleva2023temperature, zhang2021temperature} demonstrate that the temperature is the key to control the trade-off between learning instance-specific and semantic representations in contrastive loss. Similarly, temperature is also known to control the trade-off between sharper and smoother output probability distribution in the cross-entropy loss.

The main contributions of this paper are summarized as follows:
\begin{itemize}
    \item We analyze the effects of temperature scaling in an open set scenario, using 
    a number of TSs, including our novel negative cosine schedule (NegCosSch), to explore temperature modulated representation learning. We find that the proposed schedules, even simple linear schedules, demonstrate better open and closed set performance compared to the usual constant temperature baselines and possible other schedules. 
    %Even a simple linear increase performs better than the usual baseline.
    \item Our schedules can be seamlessly integrated into any existing OSR loss, such as the CE, the losses based on contrastive learning and the ARPL loss by \citet{chen2021adversarial}, without any computational overhead. We show significant performance improvements on the TinyImageNet benchmark and the SSBs.
    %\item We further integrate our method with a few existing methods for OSR and show a significant boost in their performance.  
    \item Our strategy demonstrates strong performance improvements for the tougher SSBs over the baselines for both the closed set and the open set problems. We show that our scheme achieves stronger improvements with an increased number of training classes when the task becomes more difficult for the baseline model. %improvement for fewer number of samples with more classes.

\end{itemize}

The rest of the paper is organized as follows. In Section \ref{background}, we discuss the relevant background on different losses and in Section, \ref{effect}, we discuss the effect of temperature scaling on known and unknown classes. In Section \ref{our_method}, we describe the proposed scheme. In Section \ref{results}, we discuss our results followed by related works in Section \ref{related_works} and present the concluding remarks in Section \ref{Conclusion}.

\section{Background on Losses}\label{background}

During training, the model is decomposed into two components: The first component is an encoder function $f(\cdot)$ which maps the input $x$ to a representation $z=f(x)$. The second component $h(\cdot)$ maps the representations to task specific outputs, which is either a linear classification layer if we train with the CE loss or a projection layer if we use the contrastive training. 
%The final layer essentially learns a subspace from the representations for the given objective function. 
The final outputs, also called the logits, $l=h(z)=h(f(z))$ are then given as model predictions to the loss functions. We assume for a specific problem, a model is trained for a predefined number of epochs $E$. 
We further discuss the CE and the supervised contrastive (SupCon) losses and the effects of the temperature parameter, which provide a basis for our proposed scheme. The ARPL loss is discussed in Appendix \ref{sec:ARPL}.

\subsection{Cross-entropy Loss}
For a batch of training data $\mathcal{B}=\{(x_k,y_k)\}_{k=1}^B$, the CE loss is calculated as
\begin{equation}
L_{\text{CE}} = - \frac{1}{|\mathcal{B}|}\sum_{(x_k,y_k) \in \mathcal{B}} \text{one\_hot}(y_k) \cdot \text{log}(p_k)
\end{equation}
where $p_k=\text{softmax}(l_k/\tau)$, $\text{one\_hot}(y_k)$ is the one hot encoded vector of $y_k$ and $(\cdot)$ is the dot product. The parameter $\tau>0$ is called the temperature. 

\subsection{Supervised Contrastive Loss}
For a given batch $\mathcal{B}$, the SupCon training utilizes a multi-viewed batch by taking two augmented samples of the same original sample. The multi-viewed batch $\mathcal{B}'= \{(\tilde{x_i}, \tilde{y}_i)\}_{i=1}^{2B}$, where $\tilde{x}_{2k}$ and $\tilde{x}_{2k-1}$ are two random augmentations of $x_k~~(1 \le k \le B)$, and $\tilde{y}_{2k-1} = \tilde{y}_{2k} = y_{k}$. Below we refer to 
$i$ as the anchor index from $I = \{1,...,2B\}$. 
%Here, all inputs are passed through the encoder model followed by the projection layer. 
The SupCon loss (\citet{khosla2020supervised}) is defined by:
\begin{equation}
L_{\text{SupCon}} = - \frac{1}{|I|}\sum_{i \in I} \frac{1}{|P(i)|}  \Bigg[\sum_{p \in P(i)} \text{log} \frac{\text{exp}( \text{sim}(l_i,l_p)/\tau)}{\sum_{a \in I \setminus \{i\}} \text{exp}(\text{sim}(l_i,l_a) /\tau)} \Bigg]
\label{eqn:supcon}
\end{equation}
Here, $\text{sim}(l_i,l_j)$ is the cosine similarity between $l_i$ and $l_j$. $P(i)=\{p \ne i: \tilde{y}_{i} = \tilde{y}_{p}\}$ is the set of indices in $\mathcal{B}'$ having the same label as $\tilde{y}_{i}$ and distinct from $i$. Contrastive loss, by construction, gains its strength by pushing away the representations of the negative samples (samples of other classes) and by producing compact clusters of representations for the (positive) samples of the same class.

\section{Effect of Temperature on Known and Unknown Samples}\label{effect}

%Similarly in CE loss, lower values of temperature ($\tau<1$) amplifies the relative differences among the logits resulting in a model with a sharper output probability distribution over the training classes (\citet{guo2017calibration}). On the other hand, higher values of $\tau >1$ reduces the relative differences among logits making the output probability distribution smoother. 

The SupCon loss applies hard negative mining from penalizing the harder negative samples more through the exponential function (\citet{khosla2020supervised}). The measure of hardness of a sample with respect to an anchor is determined by the scaled similarity. Therefore as a scaling factor, the temperature plays a critical role in controlling the trade-off between uniformity and semantic structure in the representation space as shown in \citet{wang2021understanding,kukleva2023temperature} for the self-supervised loss (\citet{chen2020simple}). This effect mostly translates to the supervised case except for the fact that the definition of positive and negative samples are now different.

For any given anchor index $i$, the gradient of $L_{SupCon}$ with respect to a negative logit $l_j$ can be computed as shown in the following equation.
\begin{equation}
\frac{\partial L_{SupCon}}{\partial l_j} = \frac{\partial L_{SupCon}}{\partial \text{sim}({l_i,l_j})} \times\frac{\partial \text{sim}({l_i,l_j})}{\partial l_j} = \frac{1}{\tau} [\text{softmax}_{a\in I \setminus\{i\}}(\text{sim}(l_i,l_a)/\tau)]_j \times\frac{\partial \text{sim}({l_i,l_j})}{\partial l_j}
\end{equation}
For smaller values of temperature $\tau$, as the differences in scaled similarity get amplified, the nearest negative samples receive the highest gradient (\citet{wang2021understanding}) and the model minimizes similarity to them with respect to anchor $i$. The model aggressively pushes the nearest negative samples away, leading to features that are appropriate for instance-level discrimination and distributing the embeddings over the representation space. However, the positive samples do not cluster tightly because, like the negative samples, fewer positive neighbors get priority in the loss function (Figure \ref{fig0:subfig_a}). The resulting decision boundary is sharper. The open set representations do not get closer to the known classes due to the heavy penalty of having slight dissimilarity.  

With larger $\tau$, the differences in scaled similarity diminish and the repulsive force gets distributed to more negative neighbors. The model can decrease the loss by learning the group-wise features rather than the instance discriminating features to push away easy negatives, inducing semantic structures. Due to compact clusters of within-class representations, the resulting decision boundary is smoother. However, the between-class distances also reduce as the model is less aggressive in removing the negatives. In this case a lot of open set examples get close to the known classes (Figure \ref{fig0:subfig_c}). 

Similarly in CE loss, lower values of temperature ($\tau<1$) leads to a sharper output probability distribution over the training classes (\citet{guo2017calibration}), while the higher values of $\tau >1$ makes the output probability distribution smoother. 

The value of $\tau$ is usually kept constant throughout the entire training for both losses, which is set either to a predefined value or chosen with hyperparameter tuning.

\section{Proposed Method}\label{our_method}

In this section, we formally introduce our problem, describe our proposed temperature modulation and explain how our schedules lead to learning representations useful for OSR. 

\subsection{Problem Definition}
We are given a labeled training dataset $\mathcal{D}_{tr} = \{(x_i,y_i)\} \subset \mathcal{X} \times \mathcal{Y}$, where $x_i$ is the training sample with label $y_i$. $\mathcal{X}$ is the input space and the labels of $\mathcal{D}_{tr}$ come from a closed set label space $\mathcal{Y}$, i.e. $y_i \in \mathcal{Y}, \forall i$. The total number of classes in the closed set is $C = |\mathcal{Y}|$. The test dataset $\mathcal{D}_{test} \subset \mathcal{X} \times \mathcal{Y} \cup \mathcal{O}$ consists of samples whose label space $\mathcal{Y} \cup \mathcal{O}$ is different than $\mathcal{Y}$, and $\mathcal{O} \cap \mathcal{Y} = \emptyset$. $\mathcal{O}$ is the set of unknown classes defined as the open set. The objective of OSR is to classify a test sample among the closed set classes or to flag it as belonging to an unknown class. We assume that information about the nature of unknown classes or any auxiliary samples are unavailable during training.

\begin{wrapfigure}{l}{0.62\columnwidth}
%\begin{figure}
  \centering
    \includegraphics[width=0.55\textwidth]{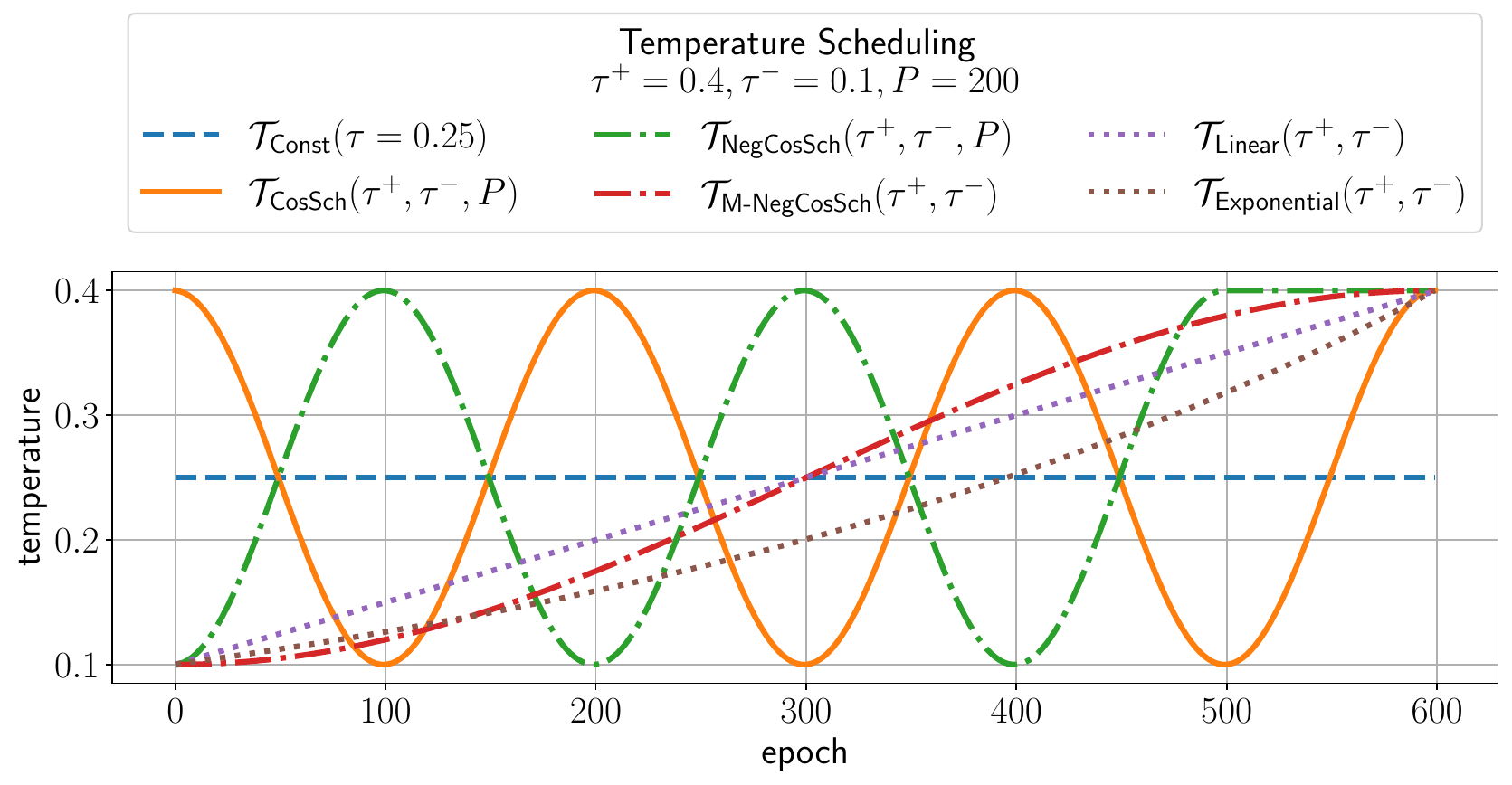}
  \caption{Different temperature schedules.}\label{fig:temp_sche}
%\end{figure}
\end{wrapfigure}

\subsection{Rationale for Temperature Scheduling}%\label{Model_training}
For a temperature dependent loss function, $\tau$ is usually kept constant throughout the training process. We denote a constant TS by $\mathcal{T}_{\text{Const}}$, defined as 
\begin{equation}
\mathcal{T}_{\text{Const}}(e;\tau) = \tau, \forall e
\end{equation}
Where $e$ is the epoch number.

For OSR, if the test representation from a novel class is too class-specific, the model easily finds its similarity to one of the known classes. On the other hand, if the feature is too instance-specific, the model is under-confident in assigning any sample to a known class. Moreover, \textit{Familiarity Hypothesis} by \citet{dietterich2022familiarity} states that most existing OSR methods flag semantic novelty from the absence of learned semantic features and it recommends to extract features for \textit{interesting content} beyond the semantic features for detecting novelty. Therefore, a model needs to capture both features -- good class-specific representations while having room for instance-level discriminating power within the class. If for the group-wise features, there is similarity between a novel sample and a known class, the instance-specific features should maintain the separation between them. A specific temperature throughout the training restricts the opportunity to explore this trade-off that may lead to learning better representations. Utilizing the effects observed in Section \ref{effect}, we propose to gradually switch between the two objectives using a generalized cosine schedule, which we describe next.

\subsection{Proposed Temperature Schedules}\label{Model_training}
%\subsubsection{Generalized Periodic Cosine Scheduling} 
Instead of a constant value, we propose to schedule the temperature (replacing $\tau$ by $\mathcal{T}(e)$) from a range $[\tau^{-},\tau^{+}]$ as the training progresses using a generalized cosine schedule, which is defined as
\begin{equation}
\mathcal{T}_{\text{GCosSch}}(e;\tau^{+},\tau^{-},P,k) = \left\{
	\begin{array}{ll}
		\tau^{-} + \frac{1}{2}(\tau^{+}-\tau^{-})(1+\text{cos}(\frac{2\pi e}{P} - k\pi)), & \mbox{if } e\le E - \frac{kP}{2}\\
		\tau^{+}, & \mbox{elsewhere } 
	\end{array}
\right.\label{gcos}
\end{equation}
%\[E'=E - \frac{kT}{2}\]
where value $k\pi$ represents the delay of cosine wave with respect to the starting epoch, $P$ is the period of the wave (Figure \ref{fig:temp_sche}) and $k$ can be from $[0,1]$. $k=0$ reduces Eq. (\ref{gcos}) to the regular cosine schedule (CosSch), proposed by \cite{kukleva2023temperature} for the self-supervised tasks on the long-tailed datasets. With CosSch, the model starts training with a higher temperature $\tau^{+}$ and goes to a lower temperature $\tau^{-}$. 
\begin{equation}
\mathcal{T}_{\text{CosSch}}(e;\tau^{+},\tau^{-},P)= \mathcal{T}_{\text{GCosSch}}(e;\tau^{+},\tau^{-},P,k=0)\label{eq:cos}
\end{equation}
\textbf{Proposed Negative Cosine Schedule.} We find that rather than using Eq. (\ref{eq:cos}), it is beneficial for the task switching if we start with a lower temperature $\tau^{-}$ and move towards a higher value $\tau^{+}$. Starting with a lower temperature, the model provides priority to fewer neighbors, learning the coarse structure of representation space, resulting in a sharper decision boundary. The open set samples remain distributed and distant from any cluster of known classes (due to heavy penalization of slight dissimilarity for the lower $\tau^-$).
As temperature increases, the model prioritizes more neighbors and gradually pulls the positive samples to its own cluster, refining on the coarse representation space. This makes within-class representations more compact and the decision boundary smoother while the core separation learned earlier is maintained. The open set samples are not pulled as tightly because their features are unknown to the model, maintaining the separation. This leads to a richer and potentially more generalized representation space for both open set and closed set performance. 

The second half cycle (decreasing from a higher to a lower temperature) facilitates the exploration by refining the model again for the instance-specific features and a smooth transition for the restart of next periodic cycle. The periodic restart can help the model to find a better nearby solution to generalize more effectively. The model settles down better if the final few epochs maintain a higher temperature rather than follow the wave (epochs 500-600 in Figure \ref{fig:temp_sche}).
Figure \ref{fig:0} illustrates the concept with UMAP projection of representation spaces for different TSs.

For $k=1$ in Eq. (\ref{gcos}), the temperature starts with a lower value and goes to a higher one, looking like a negative cosine wave, hence the name negative cosine schedule (NegCosSch).
\begin{equation}
\mathcal{T}_{\text{NegCosSch}}(e;\tau^{+},\tau^{-},P)= \mathcal{T}_{\text{GCosSch}}(e;\tau^{+},\tau^{-},P,k=1)\label{negcos}
\end{equation}
%Through experimental results, we find that NegCosSch is better for OSR than CosSch or any other value of $k$ in GCosSch.
Our experiments demonstrate that NegCosSch surpasses both CosSch and GCosSch (with other values of $k \ne 1$) in OSR performance.
Although initially aimed for OSR, NegCosSch also improves closed set classification, while being applicable to any model architecture and loss function, such as CE, SupCon and ARPL \citep{chen2021adversarial} and incurring no additional computational burden, as it only includes an epoch-dependent temperature in a loss function.

\textbf{Choice of $P$ and Other Proposed Monotonic Schedules.} We find that a single \textit{monotonic} increase with the first half cycle of NegCosSch (termed as Monotonic-NegCosSch or M-NegCosSch), where $P=2E$, is a sufficiently good TS. 
\begin{equation}
\mathcal{T}_{\text{M-NegCosSch}}(e;\tau^{+},\tau^{-}) = \tau^{-} + 0.5(\tau^{+}-\tau^{-})(1-\text{cos}(e\pi/E)); \forall e
\label{m-negcos}
\end{equation}
Even a linear or an exponential temperature increase performs better than the baseline constant temperature. The exact formulations of these appear in Appendix \ref{mono_increases}.
Otherwise, $P$ in Eq. (\ref{negcos}) can be chosen by dividing $E$ by the number of cycles. We denote periodic NegCosSch as P-NegCosSch. From ablation studies, we observe that varying $P$ does not change the performance significantly.

\textbf{Choice of $(\tau^+,\tau^-)$.}
For any good value of constant temperature $\tau$ (which can be chosen from hyperparameter tuning) in the SupCon loss, it is better to use NegCosSch by setting $\tau^+ = \tau + \Delta, \tau^- = \tau - \Delta$ (or alternatively, $\tau^+ = \tau + \Delta, \tau^- = \tau$) with the increment $\Delta \approx 0.1$ or $0.2$. 
%For example, in TinyImageNet, we find from hyperparameter tuning that $\mathcal{T}_{\text{Const}}(\tau=0.2)$ provides a good OSR performance.
For example, hyperparameter tuning on TinyImageNet provides us a high OSR performance for $\mathcal{T}_{\text{Const}}$ with $\tau=0.2$.
Therefore, $\mathcal{T}_{\text{NegCosSch}}(\tau^+ = 0.4, \tau^- = 0.1)$ or $\mathcal{T}_{\text{NegCosSch}}(\tau^+ = 0.3, \tau^- = 0.2)$ are better choices than $\mathcal{T}_{\text{Const}}(\tau=0.2)$, $\mathcal{T}_{\text{CosSch}}(\tau^+ = 0.4, \tau^- = 0.1)$ and $\mathcal{T}_{\text{CosSch}}(\tau^+= 0.3, \tau^- = 0.2)$. For the CE loss, we find $\tau^+ = 2\tau, \tau^- = \tau/2$ as a good choice because in the CE loss, the temperature scales the logits instead of similarities. 
Utilizing this, we reduce the search space of $(\tau^+,\tau^-)$.

%on the SSB experiments, where we find higher performance boost discussed in Section \ref{sec:results_SSB}. 

\textbf{Inference.} For CE loss, we use the model as is for inference. However for SupCon loss, we remove the projection layer and a linear classifier is trained for evaluation. We use the maximum logit based scoring rule for OSR score. 

\section{Results and Discussion}\label{results}
In this Section, we describe the benchmarks used for evaluating our method, the experiment settings followed by results and discussion. 

\textbf{Benchmarks.} 
%We test our method on four regular OSR benchmarks: CIFAR10, CIFAR+10, CIFAR+50, and TinyImageNet. Moreover, 
Here, we present the performance with different TSs on the TinyImageNet and the SSBs. The SSBs are defined on three fine-grained datasets: the Caltech-UCSD-Birds (CUB) (\citet{wah2011caltech}), FGVC-Aircraft (\citet{krause20133d}) and Stanford Cars (SCars) (\citet{maji2013fine}). For SSBs, the open set classes are divided into `Easy' and `Hard' splits by computing the semantic similarity (based on the labeled visual attributes) of each pair of classes, the details of which can be found in \cite{vaze2022openset}. The different difficulty levels along with more training classes make these datasets harder OSR benchmarks than the other ones. Most of the OSR research does not report results on the SSBs. Moreover, we report the performance of our NegCosSch on the CIFAR benchmarks-- CIFAR10, CIFAR+10 and CIFAR+50-- from literature with their details in the Appendix.

\textbf{Training Details.} We mostly follow the experiment settings and design choices from \citet{vaze2022openset}. 
For TinyImageNet, we use a VGG32-like model and for SSBs, we use a ResNet50 model pretrained on the places365 dataset\footnote{In spite of our efforts, we could not find the same pretrained model mentioned in \citet{vaze2022openset} online. Therefore, we use the pretrained model from \citet{zhou2017places} trained on places365, which is completely unrelated to the SSBs.}. We run each experiment with 5 random seeds and report the average results. We also include results on a vision transformer model in Appendix \ref{sec:ViT}.

We perform ablations on temperatures from $\mathbb{T}_{\text{SupCon}} = \{0.025, 0.05, 0.1, 0.2, 0.3, 0.4, 0.5\}$ for the SupCon loss and from $\mathbb{T}_{\text{CE}}=\{0.5, 1.0, 2.0\}$ for the CE loss. However following Section \ref{Model_training}, for SSBs, we fix $(\tau^+,\tau^-)$ at $(0.3,0.1)$ and $(0.4,0.1)$ for TinyImageNet with TSs and $\tau=0.2$ for $\mathcal{T}_{\text{Const}}$ in the SupCon loss. For CE loss, with the base temperature being $1.0$, we set $(\tau^+,\tau^-)$ at $(0.5,2.0)$.
Similarly, we perform an ablation on $P$ from $\{100,200,1200\}$; however, here we report results using $P=200$ for all periodic TSs and $P=2E=1200$ for M-CosSch and M-NegCosSch to ensure consistent comparison. 
Ablations on $(\tau^+,\tau^-)$, $P$ and the training details appear in the Appendix. 

\textbf{Metrics.} We report the closed set performance as a $C$-class classification using accuracy (\%), the open set performance as known-unknown detection using AUROC (\%), and the open set classification rate (OSCR \%). OSCR measures the trade-off between the open set and the closed set performance (\citet{dhamija2018reducing}). Now, we discuss the results.

\begin{table}[t]
%\begin{minipage}[h]{0.75\textwidth}
  \caption{Comparison of different TSs on CE loss. For SSBs, the OSR results are shown on `Easy/ Hard' splits. We bold the top three results for each metric and underline the best case.}
  \label{table:schedule}
  %\centering
  \scriptsize
  %\footnotesize
  \setlength{\tabcolsep}{6pt}
\begin{tabular}{lrrrrrrr}\toprule
 &Accuracy (\%) &AUROC (\%) &OSCR (\%) &Accuracy (\%) &AUROC (\%) &OSCR (\%) \\\midrule
Schedule&\multicolumn{3}{c}{\textbf{CUB}} &\multicolumn{3}{c}{\textbf{Aircraft}} \\\midrule
Constant (Baseline) &84.43 &83.55 / 74.98 &70.49 / 63.34 &90.88 &90.35 / 81.48 &82.05 / 74.25 \\
Linear decrease &81.64 &79.86 / 71.75 &65.15 / 58.59 &90.58 &89.53 / 79.7 &81.08 / 72.41 \\
Random &85.06 &85.02 / 75.54 &72.28 / 64.32 &91 &90.76 / 82.37 &82.55 / 75.17 \\
P-CosSch &84.63 &84.5 / 74.24 &71.51 / 62.93 &90.8 &90.04 / 81.81 &81.76 / 74.51 \\
M-CosSch &81.77 &79.55 / 71.4 &64.96 / 58.35 &90.62 &88.63 / 80.92 &80.35 / 73.57 \\
Logarithmic increase &85.15 &84.91 / 76.07 &72.25 / 64.82 &\textbf{91.19} &90.86 / 82.58 &82.77 / 75.47 \\
Exponential increase (ours)&\textbf{86.12} &\textbf{86.65 / 78.05} &\textbf{74.64 / \underline{67.35}} &90.88 &90.92 / 82.93 &82.54 / 75.54 \\
Linear increase (ours) &\textbf{86.22} &86.54 / \textbf{78.01} &74.58 / \textbf{67.32} &90.97 &\textbf{91.11 / \underline{83.25}} &\textbf{82.87 / 76} \\
P-NegCosSch (ours) &\textbf{\underline{86.3}} &\textbf{\underline{86.85}} / 77.6 &\textbf{\underline{74.89}} / 67.01 &\textbf{\underline{91.33}} &\textbf{\underline{91.41}} / \textbf{83.15} &\textbf{\underline{83.43 / 76.14}} \\
M-NegCosSch (ours) &\textbf{86.12} &\textbf{86.79 / \underline{78.08}} &\textbf{74.7 / 67.3} &\textbf{91.15} &\textbf{91.15} / \textbf{83.23} &\textbf{82.99 / 76} \\\midrule
&\multicolumn{3}{c}{\textbf{SCars}} &\multicolumn{3}{c}{\textbf{TinyImageNet}} \\\midrule
Constant (Baseline) &96.76 &94.03 / 84.82 &91.04 / 82.19 & 84.55&	82.85&	74.74\\
Linear decrease &96.25 &92.51 / 83.47 &89.14 / 80.51 & 84.12&	82.64	&74.28 \\
Random &97.06 &94.31 / 85.27 &91.58 / 82.85 & 83.51	&78.37&	69.88 \\
P-CosSch &96.63 &93.85 / 84.88 &90.75 / 82.14 & 84.41	&\textbf{83.12}&	74.79 \\
M-CosSch &96.27 &92.21 / 82.7 &88.84 / 79.73 & 84.19	&82.76&	74.36 \\
Logarithmic increase &97.07 &94.92 / 85.42 &92.18 / 83.03 & 84.74	&82.96&	74.91 \\
Exponential increase (ours) &\textbf{97.27} &\textbf{95.08} / 86.03 &\textbf{92.5} / 83.75 & \textbf{\underline{84.98}}	&\textbf{83.05}&	\textbf{75.15} \\
Linear increase (ours) &97.19 &\textbf{\underline{95.19}} / \textbf{86.18} &\textbf{92.55 / 83.86} & \textbf{84.9}	&\textbf{\underline{83.16}}&	\textbf{\underline{75.19}} \\
P-NegCosSch (ours) &\textbf{\underline{97.3}} &95.03 / \textbf{86.05} &92.49 / \textbf{83.81} & \textbf{84.85}	&83.02&	\textbf{75 }\\
M-NegCosSch (ours) &\textbf{97.22} &\textbf{95.18} / \textbf{\underline{86.26}} &\textbf{\underline{92.57 / 83.95}} &84.24&	82.79&	74.41 \\
\bottomrule
\end{tabular}
%\end{minipage}\hfill
%\begin{minipage}[h]{0.18\textwidth}
%\end{minipage}
\end{table}

\subsection{Ablation Study on Temperature Schedules}
Here, we compare the closed set and open set performance among different TSs, such as a random schedule, a linear decrease and an increase, exponential and logarithmic monotonic increases, periodic and monotonic CosSch and NegCosSch on CE loss. We present the results in Table \ref{table:schedule}.
For most cases, our proposed TSs, such as P-NegCosSch, M-NegCosSch, linear and exponential increases perform better than the constant baseline, CosSchs and other listed schedules in terms of all metrics. In terms of a single best result across a column, P-NegCosSch wins at a maximum number (8 out of 18) of cases and M-NegCosSch and linear win at 4 cases each. The standard deviations of these results across the trials are presented in Table \ref{table:schedule_std} in the Appendix.

\begin{table}[t]\centering
  \caption{Performance on different OSR loss functions, with and without the proposed schedules. Open set results are shown on `Easy / Hard' splits. We highlight the cases where our TS produces better results than the baseline and underline the best result for each case.}
  \label{table:SSB}
%\resizebox{ extwidth}{!}{ % use this if the table is too large
  \scriptsize
  \setlength{\tabcolsep}{4pt}
\begin{tabular}{llrrrrrrrr}\toprule
&&Accuracy (\%) &AUROC (\%) &OSCR (\%) &Accuracy (\%) &AUROC (\%) &OSCR (\%) \\\midrule
Loss& Schedule &\multicolumn{3}{c}{\textbf{CUB}} &\multicolumn{3}{c}{\textbf{Aircraft}}  \\\midrule
\multirow{3}{*}{CE (w/o LS)}& Constant &84.43 &83.55 / 74.98 &70.49 / 63.34 &90.88 &90.35 / 81.48  &82.05 / 74.25 & \\
& M-NegCosSch(\textbf{ours}) &\textbf{86.12} &\textbf{86.79} / \textbf{\underline{78.08}} &\textbf{74.7} / \textbf{\underline{67.3}} &\textbf{91.15} &\textbf{91.15 / \underline{83.23}}  &\textbf{82.99 / 76} & \\
& P-NegCosSch(\textbf{ours}) &\textbf{\underline{86.3}} &\textbf{\underline{86.85}} / \textbf{77.6} &\textbf{\underline{74.89}} / \textbf{67.01} &\textbf{\underline{91.33}} &\textbf{\underline{91.41} / 83.15}  &\textbf{\underline{83.43 / 76.14}} & \\\midrule
\multirow{3}{*}{\parbox{1.5cm}{CE + LS (\citet{vaze2022openset})}}& Constant &85.53 &85.15 / 77.44 &72.77 / 66.26 &90.73 &86.85 / 79.72 &78.84 / 72.55 \\
& M-NegCosSch(\textbf{ours}) &\textbf{\underline{86.21}} &\textbf{\underline{87.66 / 79.06}} &\textbf{\underline{75.53 / 68.23}} &\textbf{\underline{91.34}} &\textbf{\underline{88.25 / 81.19}} &\textbf{\underline{80.62 / 74.36}} \\
& P-NegCosSch(\textbf{ours}) &\textbf{86.12} &\textbf{86.43} / \textbf{78.03} &\textbf{74.36} / \textbf{67.22} &\textbf{91.1} &\textbf{87.25} / \textbf{80.03} &\textbf{79.55} / \textbf{73.17} \\\midrule
\multirow{3}{*}{\parbox{1.5cm}{SupCon (w/o LS)}}& Constant&83.43 &86.94 / 73.95 &72.42 / 61.66 &90.71 &88.78 / 81.79 &80.51 / 74.37 \\
& M-NegCosSch(\textbf{ours}) &\textbf{\underline{85.3}} &\textbf{\underline{88.14 / 75.81}} &\textbf{\underline{75.09 / 64.72}} &\textbf{\underline{91.43}} &\textbf{\underline{90.45 / 82.49}} &\textbf{\underline{82.57 / 75.51}} \\
& P-NegCosSch(\textbf{ours}) &\textbf{84.12} &\textbf{87.5} / \textbf{74.95} &\textbf{73.54} / \textbf{63.13} &90.61 &\textbf{89.27} / \textbf{81.97} &\textbf{80.96} / \textbf{74.55} \\\midrule
\multirow{3}{*}{SupCon + LS}& Constant &83.72 &86.43 / 73.69 &72.3 / 61.74 &90.05 &88.97 / 81.81 &80.11 / 73.85 \\
& M-NegCosSch(\textbf{ours}) &\textbf{\underline{85.28}} &\textbf{\underline{88.05 / 75.78}} &\textbf{\underline{74.97 / 64.63}} &\textbf{\underline{90.55}} &\textbf{\underline{89.47 / 81.85}} &\textbf{\underline{80.95 / 74.28}} \\
& P-NegCosSch(\textbf{ours}) &\textbf{84.38} &\textbf{87.26} / \textbf{75.16} &\textbf{73.5} / \textbf{63.39} &\textbf{90.43} & 88.77 / 81.78 &\textbf{80.2} / \textbf{74.08} \\\midrule
\multirow{3}{*}{\parbox{1.5cm}{ARPL (\citet{chen2021adversarial})}}& Constant &
85.8 &86.93 / 79.7 & 78.64 / 73.36 & 90.88 &90.75 / 81.77 &85.98 / 78.26 \\
& M-NegCosSch(\textbf{ours}) & \textbf{86.47} &\textbf{\underline{87.6 / 80.53}} &\textbf{\underline{79.65 / 74.41}} &\textbf{\underline{91.26}} &\textbf{\underline{91.55}} / \textbf{82.01} &\textbf{\underline{86.65 / 78.39}} \\
& P-NegCosSch(\textbf{ours}) & \textbf{\underline{86.51}} &\textbf{87.57} / \textbf{80.12} &\textbf{79.61} / \textbf{74.05} &\textbf{90.98} &\textbf{91.52} / \textbf{\underline{82.02}} &\textbf{86.55} / \textbf{78.38} \\\midrule
&&\multicolumn{3}{c}{\textbf{SCars}} &\multicolumn{3}{c}{\textbf{TinyImageNet}}  \\\midrule
\multirow{3}{*}{CE (w/o LS)}&Constant &96.76 &94.03 / 84.82 &91.04 / 82.19 &81.95  &78.6  &69.22 \\
& M-NegCosSch(\textbf{ours}) &\textbf{97.22} &\textbf{\underline{95.18 / 86.26}} &\textbf{\underline{92.57 / 83.95}} &\textbf{81.98} & \textbf{\underline{79.21}}  &\textbf{69.84} \\
& P-NegCosSch(\textbf{ours}) &\textbf{\underline{97.3}} &\textbf{95.03} / \textbf{86.05} &\textbf{92.49} / \textbf{83.81} &\textbf{\underline{82.23}} &\textbf{79.05}  &\textbf{\underline{69.91}} \\\midrule
\multirow{3}{*}{CE + LS}&Constant &97.05 &94.67 / 84.35 &91.95 / 82.02 &84.55  &82.85 &74.74 \\
& M-NegCosSch(\textbf{ours}) &\textbf{\underline{97.23}} &\textbf{\underline{95 / 85.06}} &\textbf{\underline{92.42 / 82.83}} &84.24 &82.79 &74.41 \\
& P-NegCosSch(\textbf{ours}) &\textbf{\underline{97.23}} &\textbf{94.82} / \textbf{84.54} &\textbf{92.24} / \textbf{82.31} &\textbf{\underline{84.85}} &\textbf{\underline{83.02}} &\textbf{\underline{75}} \\\midrule
\multirow{3}{*}{\parbox{1.5cm}{SupCon (w/o LS)}}&Constant &96.58 &92.99 / 82.8 &89.92 / 80.12 &85.37 &82.87 &70.61 \\
& M-NegCosSch(\textbf{ours}) &\textbf{\underline{96.79}} &\textbf{\underline{93.57}} / 82.76 &\textbf{\underline{90.66}} / \textbf{80.26} &\textbf{\underline{85.4}} &\textbf{\underline{83.21}} &\textbf{\underline{70.98}} \\
& P-NegCosSch(\textbf{ours}) &\textbf{96.68} &\textbf{93.32} / \textbf{\underline{83.16}} &\textbf{90.31} / \textbf{\underline{80.53}} &85.18  &\textbf{83.09}  &\textbf{70.71} \\\midrule
\multirow{3}{*}{SupCon + LS}& Constant &96.6 &93.03 / \underline{83.32} &89.95 / 80.63 &85.18  &82.65  &70.31 \\
& M-NegCosSch(\textbf{ours}) &\textbf{\underline{96.84}} &\textbf{\underline{93.58}} / 83.15 &\textbf{\underline{90.69}} / 80.62 &\textbf{\underline{85.57}}  &\textbf{\underline{83.11}}  &\textbf{\underline{71.04}} \\
& P-NegCosSch(\textbf{ours}) &\textbf{96.69} &\textbf{93.45} / 83.29 &\textbf{90.43} / \textbf{\underline{80.66}} &\textbf{85.23}  &\textbf{83.05}  &\textbf{70.72} \\\midrule
\multirow{3}{*}{ARPL}& Constant & \underline{97.37} &95.22 / 85.89 &93.46 / 84.7 &\underline{85.02} &83 &74.89 \\
& M-NegCosSch(\textbf{ours}) &97.29 &\textbf{\underline{95.27 / 86.03}} &\textbf{\underline{93.48 / 84.82}} &84.83 &\textbf{83.07} &\textbf{75.03} \\
& P-NegCosSch(\textbf{ours}) &97.21 &\textbf{95.25} / 85.71 &\textbf{93.47} / 84.52 & 85  &\textbf{\underline{83.12}} &\textbf{\underline{75.07}} \\
\bottomrule
\end{tabular}
\end{table}

\subsection{Performance of our Temperature Schedules across Various Loss Functions}\label{sec:results_SSB}
Here, we report the performance on different OSR loss functions by including and without our NegCosSch in Table \ref{table:SSB}. We implement the ARPL loss, the CE baseline by \citet{vaze2022openset} and the widely implemented SupCon loss in the recent OSR literature \citep{xu2023contrastive,zhou2024contrastive}. As label smoothing (LS) has shown performance improvements, we experiment both with and without uniform LS for the CE and SupCon losses. Here, we do not optimize performance for the LS coefficient and temperatures but use a fixed set of hyperparameters for consistency.

We observe that including the proposed schedules (P-NegCosSch or M-NegCosSch) in any OSR loss function improves performances both for the closed set and open set problems over the corresponding constant temperature baseline for all cases except for two, such as for Aircraft on SupCon loss including LS and P-NegCosSch, and for TinyImageNet on CE loss including LS and M-NegCosSch. Our NegCosSch provides performance boost for up to 1.87\% of accuracy, up to 3.3\%/3.1\% of open set AUROC in the `Easy'/`Hard' splits and up to 4.4\%/3.96\% of OSCR. This amount of performance boost comes without any additional computational cost. Between our two schedules, the M-NegCosSch performs better in most cases than the periodic one.

Moreover, our method can be used together with LS to further boost the performance in a few cases.
Even for cases where LS does not help the open set performance (for the Stanford cars -`Hard' split and Aircraft with CE loss), our NegCosSch outperforms the baselines. Overall for SSBs, the CE loss performs better than SupCon loss. We believe that our scheme, in principle, can improve other OSR methods, such as the method by \citet{jia2024revealing}.

\begin{figure}[t]
  \centering
  \begin{subfigure}[b]{0.325\textwidth} 
    \centering
    \includegraphics[width=\textwidth]{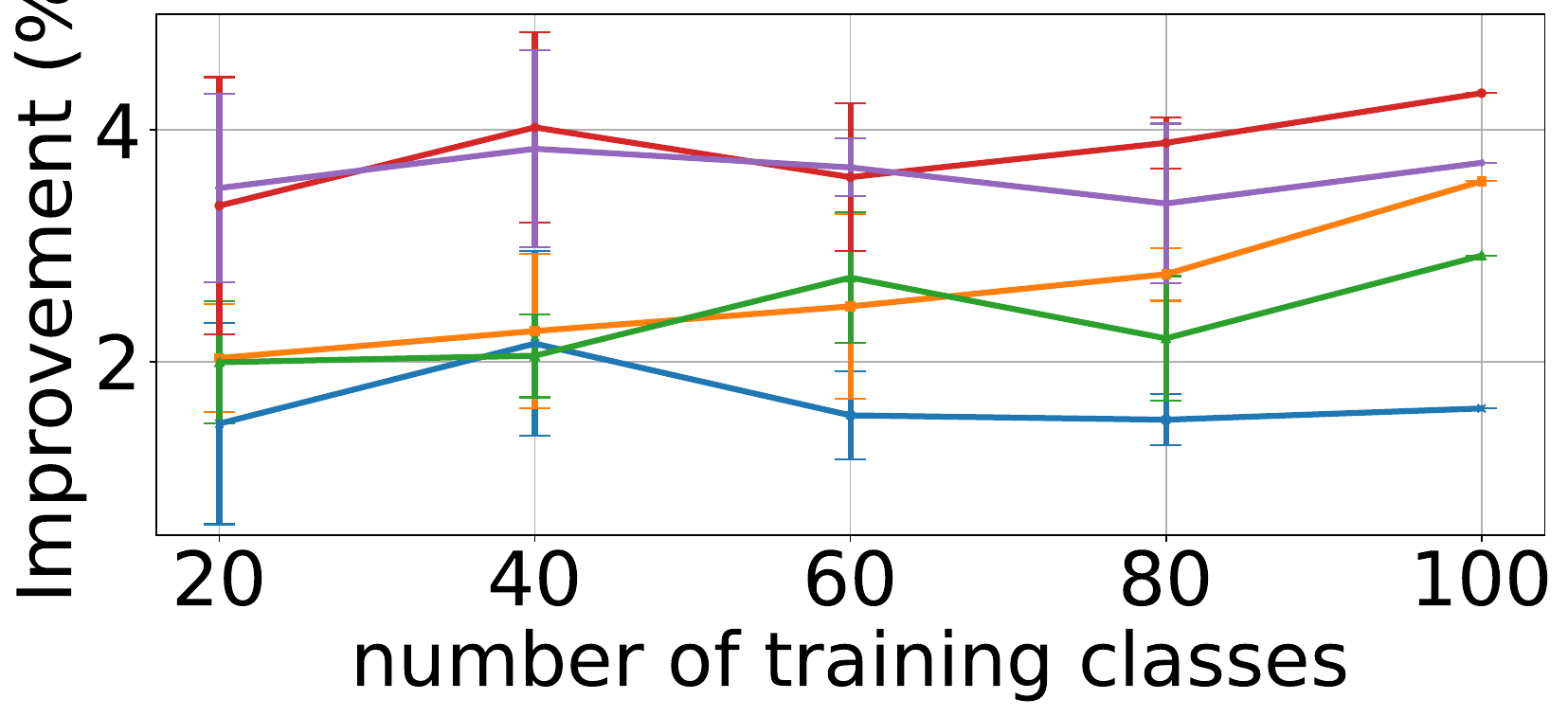}
    \caption{CUB} % 
    \label{fig2:subfig_a} 
  \end{subfigure}
  \begin{subfigure}[b]{0.325\textwidth} 
    \centering
    \includegraphics[width=\textwidth]{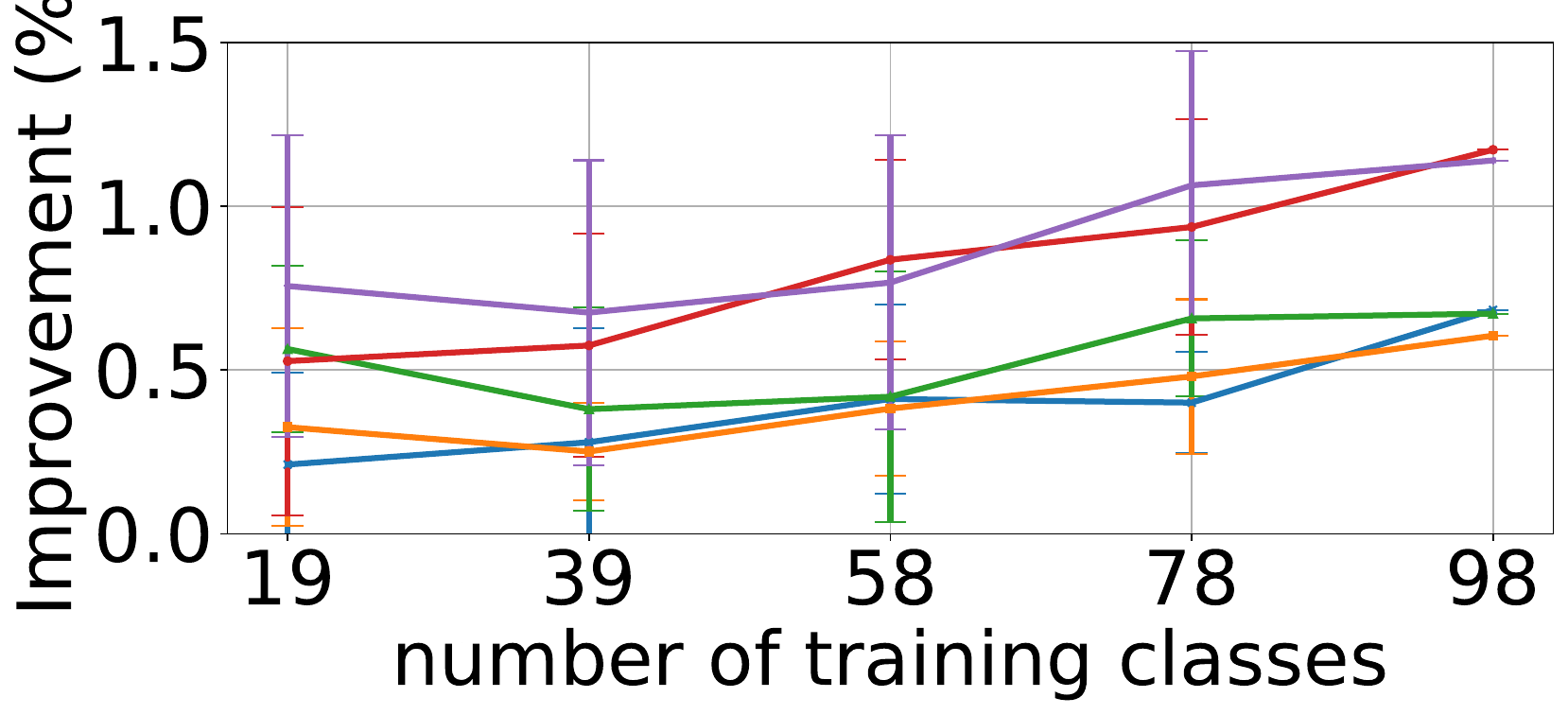}
    \caption{SCars} % 
    \label{fig2:subfig_b} 
  \end{subfigure}
  \begin{subfigure}[b]{0.325\textwidth} 
    \centering
    \includegraphics[width=\textwidth]{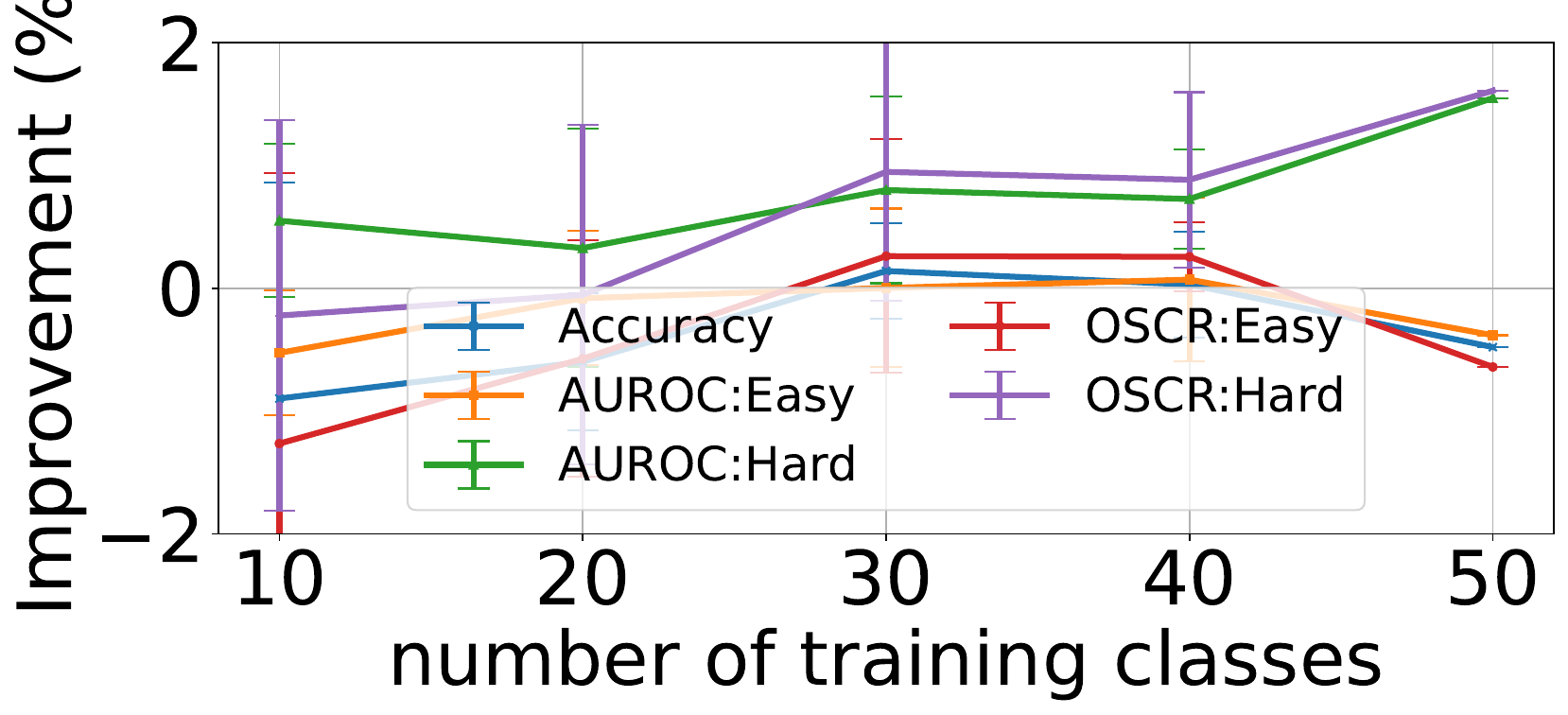}
    \caption{FGVC-Aircraft} % 
    \label{fig2:subfig_c} 
  \end{subfigure}
  \caption{Effect on performance improvement for our proposed schedule over the baselines with varying number of training classes. 
  Increasing the number of training classes tends to yield greater improvements in OSCR across all datasets, along with significant improvements in AUROC and accuracy, with the effect being most pronounced for CUB and least for FGVC-Aircraft. 
  %Overall upward improvement trend for most cases indicates that the proposed TS benefits more with more training classes.
  Error bars represent the standard deviations across trials with random training classes.}\label{fig:2}
\end{figure}

\subsection{Proposed schedule is more beneficial with more training classes}
To show the strength of our NegCosSch with increased number of training classes, we train models on CE loss with $\{20\%, 40\%, 60\%, 80\%, 100\%\}$ of the randomly chosen training classes for the SSBs without changing the open set. In Figure \ref{fig:2}, we plot the improvement of a metric $m$ over the corresponding baseline, defined as:
%\begin{equation}
$\text{improvement} = m[\mathcal{T}_{\text{NegCosSch}}(\tau^+,\tau^-)] - \max_{\tau \in \mathbb{T}_{CE}} \{m[\mathcal{T}_{\text{Const}}(\tau)]\}\notag$. 
%\end{equation}
We observe an overall upward improvement trend with more training classes in most cases for both the `Easy' and `Hard' OSR splits. 
The negative values in Figure \ref{fig2:subfig_c} is due to the fact that we measure improvement over the maximum score of three baselines.  
With more training classes, the task becomes harder for the base model, which is observed by the performance decline. Nonetheless, our schedule gains higher improvement with more training classes.

\section{Related Works}\label{related_works}
\textbf{Open Set Recognition.} Since the introduction of OSR problem, it has received a significant interest in the research community. Most of the research attempts can be summarized into several common categories, some of which are discussed in Section \ref{sec:intro}. Besides the use of generative models, input or mani-fold mix-up, other works add auxiliary samples with different strong augmentations for training models \citep{wang2025backmix, jiang2023openmix+, jia2024revealing, xuinformed}.
Another huge group of research depends on training an additional model with a secondary objective function (\cite{oza2019c2ae, sun2020open, perera2020generative, yoshihashi2019classification, zhang2020hybrid, jia2024revealing, zhou2024contrastive}). However, training generative models or a secondary model is tough for larger benchmarks as it requires significant computation overhead.

%The early ones are based on extreme value theory that modify the softmax function to model the novel classes (\citet{vignotto2018extreme, bendale2016towards}). Several works use generative models to generate synthetic unknown or counter-factual samples (\citet{ge2017generative, neal2018open, chen2021adversarial, moon2022difficulty}), while other add auxiliary samples in the dataset with strategies, such as the sample mix-up (\citet{chen2021adversarial,xu2023contrastive,bahavan2025sphor, li2024all}), manifold mix-up (\citet{zhou2021learning}), or by adding extra augmented samples (\citet{wang2025backmix, jiang2023openmix+, jia2024revealing, xuinformed}). 

Another set of methods either construct a different loss function (\citet{chen2021adversarial, chen2020learning, wang2022openauc}) or add regularization to bound the open set risks (\citet{zhou2021learning, lu2022pmal, yang2024icausalosr}). For example, the method by \citet{zhou2021learning} learns additional place-holders for the novel classes. 
Methods by \citet{chen2021adversarial, chen2020learning} learn the reciprocal points of known classes representing the `otherness' corresponding to each class. These methods try to create additional empty regions in the representation space hoping that open set representations lie in those regions.  
The new baseline by \citet{vaze2022openset} with well-trained closed set classifiers has triggered the OSR research for better representation learning schemes. For example, methods by \citet{xu2023contrastive, xu2024know, li2024all, bahavan2025sphor, li2025unlocking} train models using the contrastive loss with regularization and heavy augmentations (\citet{wang2025backmix, jiang2023openmix+, jia2024revealing}). The method by \citet{wang2024exploring} trains multiple experts for extracting diverse representations, and \citet{yang2024dynamic} proposes an open set self-learning framework, which adapts the model according to the test data assuming that it is available. Furthermore, several prior works have focused on developing fine-grained OSR methods \citep{lang2024coarse, bao2023latent, sun2023hierarchical}.

\textbf{Temperature Scaling.}
Temperature scaling in the CE loss plays a crucial role in knowledge distillation (\citet{hinton2015distilling}), model calibration (\cite{guo2017calibration}) and so on. 
Since the contrastive loss has become popular for many tasks (\citet{chen2020simple, khosla2020supervised}), studies have aimed to understand its behavior. Recently, methods by \citet{jeong2024simple, qiu2024cool} utilize temperature cool-down in language models.

\section{Conclusion}\label{Conclusion}
We develop novel temperature schedules, which can be folded into any existing OSR loss function, such as cross-entropy, contrastive or ARPL without any computational overhead. We find that starting with a lower temperature and moving towards a higher temperature results in making tighter representation clusters for the closed set classes, while the representations of the open set examples remain more distant. This process is more effective than using a fixed temperature or the opposite schedule. 
%We demonstrate improvement of open set performance for some of the well-known OSR methods on the regular benchmarks when we include our temperature scheduling. 
Our proposed schedules demonstrate strong performance improvements on the regular and the tougher semantic shift benchmarks for both closed set and open set problems for some of the well-known OSR loss functions, even on top of label smoothing. The benefit of our scheme can be better realized with a larger number of training classes.

%\subsubsection*{Author Contributions}
%If you'd like to, you may include  a section for author contributions as is done in many journals. This is optional and at the discretion of the authors.

%\subsubsection*{Acknowledgments}
%Use unnumbered third level headings for the acknowledgments. All acknowledgments, including those to funding agencies, go at the end of the paper.

\section{Reproducibility Statement}
Our implementation adheres rigorously to the benchmarks, i.e., the set of known- unknown splits defined in the standard OSR literature, such as in \citet{vaze2022openset}. For consistent comparison, we use the same experiment settings and design choices in model architecture and hyperparameters, the details of which can be found in Appendix \ref{training_details}. 
Detailed information on the hardware and software utilized is provided in Appendix \ref{implementation}. 
Project codes will be available. 
%are available at: \href{https://anonymous.4open.science/r/NegCosSch-4516/}{https://anonymous.4open.science/r/NegCosSch-4516/}. 

\bibliography{OSR_bib}
\bibliographystyle{iclr2026_conference}

\newpage
\appendix

\hfil{\LARGE Supplementary Materials}\hfil

%gradient of SupCon and CE
%SSB benchmark: system of categorization
%ProtoCon
%FGVC works?

\section{Ethics Statement}
Open set recognition is crucial for enhancing safety and reliability in machine learning systems operating in changing environments by detecting novel patterns. For instance, all categories of interest may not be represented in the training set due to their rarity or new categories may emerge due to dynamic nature. The capability of a deep model of knowing what it doesn't know enhances trust across various critical applications.

The solution for the OSR problem is yet to be improved, especially for larger datasets. Their performance depends on the semantic closeness between the known and unknown classes. Hence, the methods cannot be solely relied upon in deployment. For example, an over-sensitive OSR system can lead to a high false alarm rate.

\section{Monotonic Temperature Schedules}\label{mono_increases}
Different monotonic TSs-- such as the linear, exponential, and logarithmic increases, as well as the linear decrease, over the range $[\tau^{-},\tau^{+}]$-- are listed below. Similar to our negative cosine TS, the linear and exponential schedules also start with a lower value of $\tau$ and gradually switches the task with a higher value. For a random TS, we pick a random temperature from $[\tau^-, \tau^+]$ at each epoch.
\begin{eqnarray}
\mathcal{T}_{\text{linear}}(e;\tau^+,\tau^-) = \tau^- + \frac{e}{E}(\tau^+ - \tau^-) \\
\mathcal{T}_{\text{exponential}}(e;\tau^+,\tau^-) = \tau^- \times (\frac{\tau^+}{\tau^-})^{e/E} \\
\mathcal{T}_{\text{logarithmic}}(e;\tau^+,\tau^-) = \tau^- + (\tau^+ - \tau^-) \times \frac{\text{log}(e)}{\text{log}(E)} \\
\mathcal{T}_{\text{linear-decrease}}(e;\tau^+,\tau^-) = \tau^+ - \frac{e}{E}(\tau^+ - \tau^-) 
\end{eqnarray}

\section{Gradients of Loss Functions}
\subsection{Gradient of SupCon Loss}
For any sample $i \in I$, the gradient of $L_{\text{SupCon}}$ in Eq. (\ref{eqn:supcon}) with respect to a negative logit $l_j$
%\begin{eqnarray}
\begin{flalign}
%&= \frac{\partial L_{SupCon}}{\partial \text{sim}({l_i,l_j})} \times\frac{\partial \text{sim}({l_i,l_j})}{\partial l_j} \notag\\
\frac{\partial L_{SupCon}}{\partial l_j} &=\frac{\partial }{\partial \text{sim}({l_i,l_j})} \Bigg[ \frac{1}{|P(i)|} \sum_{p \in P(i)}\Big( -\frac{1}{\tau}\text{sim}({l_i,l_p}) + \text{log}\sum_{a \in I \setminus \{i\}} \text{exp}(\text{sim}(l_i,l_a) /\tau) \Big)\Bigg]\notag\\ 
&~~~~~~~~~~~~~~~~~~~~~~~~~ \times\frac{\partial \text{sim}({l_i,l_j})}{\partial l_j}\notag\\
&= \frac{1}{|P(i)|} \sum_{p \in P(i)} \frac{\frac{1}{\tau}\text{exp}(\text{sim}(l_i,l_j) /\tau)}{\sum_{a \in I \setminus \{i\}} \text{exp}(\text{sim}(l_i,l_a) /\tau)} \times\frac{\partial \text{sim}({l_i,l_j})}{\partial l_j}\notag\\
&= \frac{1}{\tau} [\text{softmax}_{a\in I \setminus\{i\}}(\text{sim}(l_i,l_a)/\tau)]_j \times\frac{\partial \text{sim}({l_i,l_j})}{\partial l_j} \notag
\end{flalign}
%\end{eqnarray}

We already discussed this in Section \ref{effect}. Similarly, the gradient of $L_{\text{SupCon}}$ with respect to a positive logit $l_j$
\begin{equation}
\frac{\partial L_{SupCon}}{\partial l_j} = \frac{1}{\tau |P(i)|} \Big( |P(i)|[\text{softmax}_{a\in I \setminus\{i\}}(\text{sim}(l_i,l_a)/\tau)]_j -1 \Big) \times\frac{\partial \text{sim}({l_i,l_j})}{\partial l_j}\notag
\end{equation}

%\item The value of softmax can be from $\frac{1}{|I \setminus\{i\}|}$ to 1\\
%\item when $\tau \rightarrow \infty$, softmax term $\rightarrow \frac{1}{|I \setminus\{i\}|}$\\
%$\Rightarrow \frac{\partial L_{SupCon}}{\partial l_j} \rightarrow \frac{1}{\tau |P(i)|} \Big[ |P(i)|\times softmax(.) -1 \Big] \times\frac{\partial \text{sim}({l_i,l_j})}{\partial l_j} < 0$\\
To push the gradient towards 0, value of the softmax function should approach towards $\frac{1}{|P(i)|}$. For large value of $\tau$, this is possible when all negatives are far away than the positives, to have minimum effects in the denominator of softmax function. Moreover, the differences of scaled similarities between anchor and the positives diminish, inducing the group-wise features.

\subsection{Gradient of CE Loss}
The gradient of $L_{CE}$ with respect to the logit of output node $j$ corresponding to the true label  of sample $k$,
\begin{equation}
\frac{\partial L_{CE}}{\partial l_{k,j}} = - \frac{\partial}{\partial l_{k,j}} \text{log} \frac{\text{exp}(l_{k,j}/\tau)}{\sum_i \text{exp}(l_{k,i}/\tau)}
= \frac{1}{\tau} [\text{softmax}_k(l_k/\tau)_j -1] \notag
\end{equation}
For small $\tau$, the differences of scaled logits will be amplified, and the softmax will approach towards an indicator function. The same softmax value in this gradient term computes the probability for the output node of the true class which will approach towards 1.0. Therefore, the resulting probability distribution is sharper. For large $\tau$, the differences of scaled logits will diminish, and the softmax will approach towards $1/C$, making the resulting probability distribution smoother.

\section{Training Details}\label{training_details}
\textbf{Benchmarks.} There are 5 known-unknown random splits defined in the regular OSR benchmarks. We consider 4 regular benchmarks, such as the CIFAR10, CIFAR+10. CIFAR+50 and the TinyImageNet benchmarks.
The CIFAR10 benchmark has 6 closed set classes and 4 open set classes, whereas CIFAR+10 and CIFAR+50 have 4 closed set classes from the CIFAR10 dataset and 10 and 50 open set classes from CIFAR100 dataset respectively. TinyImageNet has 20 training classes and 180 closed set classes. The SSBs are defined with 50\% of the classes in training and rest 50\% classes are divided into `Easy', `Medium' and `Hard' splits. Similar to \citet{vaze2022openset}, we combine the `Medium' and `Hard' splits to report as the `Hard' split in our paper. For each split, we run each experiment with 5 different random seeds and we report average results. 
%The random seeds used are 0, 1, 42, 47 and 107.

\textbf{Model Architecture.} We follow experimental settings similar to the existing literature, such as in \citet{vaze2022openset}. For the regular benchmarks, we train VGG32-like models from scratch and for the SSBs, we train ResNet50 models pretrained on the places365 dataset for a supervised task. The feature dimensions are 128 and 2048 for the regular benchmarks and the SSBs respectively. The linear projection layer for SupCon training has the same number of input and output nodes as the feature dimension.

\textbf{Hyperparameters.} We train all models for 600 epochs with the SGD optimizer with a momentum of 0.9 and a weight decay of $10^{-4}$. We use a cosine learning rate scheduler with warm-ups and 2 restarts at the 200-th and 400-th epoch. The initial learning rate is set to 0.1 for the CIFAR benchmarks and 0.001 for the SSBs. For TinyImageNet, it is set to 0.01 for the CE loss and we tune it to 0.05 for the SupCon loss.
Rand-Augment is used for data augmentations in all cases.
$P$ is chosen as 200 everywhere expect for the CIFAR+10 and CIFAR+50 benchmarks,
where it is set to 100. Batch size is set to 128 for the regular benchmarks. For the SSBs, the batch size is set to 12 as only this amount can be accommodated in our single GPU for each experiment in the SupCon training. 
The images are resized to $32\times32$, $64\times 64$, and $448\times448$ respectively for the CIFAR, the TinyImageNet benchmarks and the SSBs.

\textbf{Label Smoothing.} For TinyImageNet and the SSB datasets, we report results both including and without uniform LS as LS has shown improvements for these datasets. For the CE loss, we choose the LS coefficients from \cite{vaze2022openset}.
For LS in SupCon loss, we implement the following function instead of (\ref{eqn:supcon}):
\begin{equation}
L_{\text{SC,LS}} = -\frac{1}{|I|} \sum_{i \in I} \sum_{j \in I\setminus\{i\}} \frac{1}{N_i(\alpha)}[(1 - \alpha) \mathbf{1}_{y_i=y_j} + \frac{\alpha}{C-1} \mathbf{1}_{y_i\ne y_j}] \Bigg[ \text{log} \frac{\text{exp}( \text{sim}(z_i,z_j)/\tau)}{\sum_{a \in I \setminus \{i\}} \text{exp}(\text{sim}(z_i,z_a) /\tau)} \Bigg]
\label{eqn:supcon_LS}
\end{equation}
with $N_i(\alpha)=\sum_{k \in I\setminus\{i\}} [(1 - \alpha) \mathbf{1}_{y_i=y_k} + \frac{\alpha}{C-1} \mathbf{1}_{y_i\ne y_k}]$, where $\alpha$ is the smoothing coefficient and $\mathbf{1}$ is the indicator function.
For contrastive loss, we tune $\alpha$ from $\{0.1,0.2,0.3\}$; however, we use $\alpha=0.2$ in Table \ref{table:SSB} for consistent comparison.

\textbf{Details for the UMAPs in Figure 1.} We randomly choose 10 training classes from the defined closed set of the CUB benchmark and keep the open set as it is. We train models with constant temperatures of $0.5, 1.0, 2.0$, $\mathcal{T}_{\text{CosSch}}(\tau^+=2.0,\tau^-=0.5)$ and our $\mathcal{T}_{\text{NegCosSch}}(\tau^+=2.0,\tau^-=0.5)$ with CE loss and without LS. To show the training progress in our method, we plot the features at the beginning, the middle and the end of the last scheduling period starting at epoch 400.
We standardize the features by subtracting the mean and scaling them to unit variance before applying UMAP transformation. For clear visualization, we plot features of all the closed set samples and 10\% random open set samples.

\section{Discussions on the Adversarial Reciprocal Point Learning}\label{sec:ARPL}
The adversarial reciprocal point learning (ARPL) method \citep{chen2021adversarial} defines a reciprocal point for each category $c$, denoted by $r_c$, which is regarded as the latent representation of the `otherness' corresponding to each class. The reciprocal points $\{r_c\}_{c=1}^C$ are learnable parameters. 
Given a logit $l_i=f(x_i)$ and a reciprocal point $r_c$, their distance $d(l_i, r_c)$ is calculated by combining the Euclidean distance and the dot product as the following:
\begin{eqnarray}
d_e(l_i, r_c) =\frac{1}{D} ||l_i - r_c||^2_2\notag~,~~~d_d(l_i, r_c) = l_i \cdot r_c\notag\\
d(l_i, r_c) = d_e(l_i, r_c ) -d_d(l_i, r_c)\notag
\end{eqnarray}

\begin{table}[t]
  \caption{Performance standard deviation of different TSs on CE loss across various seeds.}
  \label{table:schedule_std}
  %\centering
  \scriptsize
  %\footnotesize
  \setlength{\tabcolsep}{8pt}
\begin{tabular}{lrrrrrrr}\toprule
&Accuracy (\%) &AUROC (\%) &OSCR (\%) &Accuracy (\%) &AUROC (\%) &OSCR (\%) \\\midrule
TS &\multicolumn{3}{c}{\textbf{CUB}} &\multicolumn{3}{c}{\textbf{Aircraft}} \\\midrule
Const. (Baseline) &0.2 &0.26 / 0.26 &0.41 / 0.37 &0.39 &0.77 / 0.52 &0.42 / 0.47 \\
Linear decrease &0.78 &0.35 / 1.02 &0.76 / 1.03 &0.29 &0.42 / 0.53 &0.46 / 0.56 \\
Random &0.64 &0.23 / 0.44 &0.67 / 0.74 &0.21 &0.27 / 0.75 &0.25 / 0.75 \\
P-CosSch &0.42 &0.29 / 0.52 &0.6 / 0.7 &0.31 &0.74 / 0.27 &0.85 / 0.36 \\
M-CosSch &0.28 &0.26 / 0.66 &0.06 / 0.63 &0.32 &0.82 / 1.05 &0.85 / 1.14 \\
Logarithmic increase &0.12 &0.39 / 0.48 &0.39 / 0.46 &0.32 &0.34 / 0.49 &0.53 / 0.45 \\
Exponential increase (ours) &0.2 &0.28 / 0.18 &0.29 / 0.15 &0.27 &0.26 / 0.54 &0.12 / 0.54 \\
Linear increase (ours) &0.31 &0.21 / 0.19 &0.25 / 0.15 &0.29 &0.12 / 0.56 &0.23 / 0.73 \\
P-NegCosSch (ours)&0.47 &0.38 / 0.23 &0.34 / 0.4 &0.34 &0.25 / 0.43 &0.26 / 0.5 \\
M-NegCosSch (ours)&0.27 &0.44 / 0.28 &0.6 / 0.45 &0.31 &0.15 / 0.47 &0.26 / 0.3 \\\midrule
&\multicolumn{3}{c}{\textbf{SCars}} &\multicolumn{3}{c}{\textbf{TinyImageNet}} \\\midrule
Const. (Baseline) &0.1 &0.52 / 0.49 &0.51 / 0.52 & 0.25 &	0.24 &	0.21 \\
Linear decrease &0.29 &0.76 / 0.77 &0.9 / 0.89 & 0.18 &	0.25 &	0.21 \\
Random &0.1 &0.5 / 0.79 &0.54 / 0.77 & 0.19 &	1.23 &	1.1 \\
P-CosSch &0.08 &0.56 / 0.98 &0.55 / 0.96 & 0.16 &	0.34 &	0.3 \\
M-CosSch &0.13 &0.69 / 0.62 &0.61 / 0.59 & 0.1 &	0.32 &	0.28\\
Logarithmic increase &0.15 &0.4 / 0.56 &0.41 / 0.67 & 0.2 &	0.28 &	0.33 \\
Exponential increase  (ours)&0.1 &0.31 / 0.57 &0.3 / 0.63 & 0.12 &	0.2 &	0.22\\
Linear increase  (ours)&0.1 &0.23 / 0.53 &0.27 / 0.58 & 0.24 &	0.32 &	0.37 \\
P-NegCosSch (ours) &0.1 &0.29 / 0.62 &0.21 / 0.59 & 0.29 &	0.37 &	0.36\\
M-NegCosSch (ours) &0.09 &0.22 / 0.51 &0.26 / 0.55 & 0.15 &	0.22 &	0.19 \\
\bottomrule
\end{tabular}
\end{table}

$D$ is the number of feature dimension in $l$. The final classification probability is calculated as:
\begin{equation}
p(y_i = c|x_i, f, \{r_c\}_{m=1}^C) = \frac{\text{exp}(d(l_i, r_c)/\tau)}{\sum_{m=1}^C \text{exp}(d(l_i, r_m)/\tau) }\notag
\end{equation}
The total loss is calculated as
\begin{equation}
L_{\text{ARPL}}=-\text{log}~p(y_i = c|x_i, f, \{r_c\}_{c=1}^C) + \lambda~\text{max}(d_e(l_i, r_c) - R, 0)
\end{equation}
We observe that the distance is also scaled with the temperature parameter ($\tau$) in this loss. $\lambda$ is to adjust the trade-off between the two loss terms and is set to 0.1 and, $R$ is the learnable margin parameter.

%\subsection{reporting strategy or hyperparams for each row in tables/columns}
%hypeparameters for Table \ref{table:SSB}. A LS coefficient of 0.3 and 0.2 are used respectively for the CE and the SupCon loss. We report results on 1.0 and 0.1 as the constant temperatures and for the temperature scheduling, $\text{NegCosSch}(\tau^{+}-2.0,\tau^{-}=0.5,P=200)$ and $\text{NegCosSch}(\tau^{+}=0.3,\tau^{-}=0.1,P=200)$ are used respectively for the CE and the SupCon loss. 

%\textbf{Other OSR Methods:} ConOSR ARPL implementation. While implementing ConOSR we used mean instead of sum

\section{Additional Results}

Here, we discuss the performance variability of the proposed TSs, ablation studies on $(\tau^+,\tau^-),P$, and $k$, some results on a vision transformer model, and performance on the CIFAR benchmarks.

\subsection{Performance Variability}
Here, we present the standard deviations of performance metrics across trials with 5 different seeds. Tables \ref{table:schedule_std} and \ref{table:SSB_std} present the standard deviations for the performance results reported in Tables \ref{table:schedule} and \ref{table:SSB}, respectively. The proposed TSs demonstrate either better or similar standard deviation compared to the baseline and the other TSs (presented in Table \ref{table:schedule_std}) and for all losses (presented in Table \ref{table:SSB_std}) considering the significant performance boost achieved by our proposed ones.

\begin{table}[t]\centering
  \caption{Performance standard deviation of constant baseline and our NegCosSch on different losses across various seeds.}
  \label{table:SSB_std}
%\resizebox{ extwidth}{!}{ % use this if the table is too large
  \scriptsize
  \setlength{\tabcolsep}{4pt}
\begin{tabular}{llrrrrrrrr}\toprule
& &Accuracy (\%) &AUROC (\%) &OSCR (\%) &Accuracy (\%) &AUROC (\%) &OSCR (\%) \\\midrule
Loss & Schedule &\multicolumn{3}{c}{\textbf{CUB}} &\multicolumn{3}{c}{\textbf{Aircraft}}  \\\midrule
\multirow{3}{*}{CE (w/o LS)}& Constant&0.2 &0.26 / 0.26 &0.41 / 0.37 &0.39 &0.77 / 0.52 &0.42 / 0.47 \\
& M-NegCosSch(\textbf{ours}) &0.27 &0.44 / 0.28 &0.6 / 0.45 &0.31 &0.15 / 0.47 &0.26 / 0.3 \\
& P-NegCosSch(\textbf{ours}) &0.47 &0.38 / 0.23 &0.34 / 0.4 &0.34 &0.25 / 0.43 &0.26 / 0.5 \\\midrule
\multirow{3}{*}{\parbox{1.5cm}{CE + LS (\citet{vaze2022openset})}} &Constant&0.14 &0.58 / 0.34 &0.4 / 0.23 &0.16 &0.27 / 0.54 &0.19 / 0.49 \\
& M-NegCosSch(\textbf{ours}) &0.34 &0.59 / 0.36 &0.59 / 0.38 &0.26 &0.58 / 0.29 &0.54 / 0.18 \\
& P-NegCosSch(\textbf{ours}) &0.22 &0.77 / 0.31 &0.66 / 0.41 &0.25 &0.83 / 0.35 &0.77 / 0.22 \\\midrule
\multirow{3}{*}{\parbox{1.5cm}{SupCon (w/o LS)}} & Constant&0.29 &0.23 / 0.29 &0.35 / 0.23 &0.58 &1.79 / 0.53 &1.99 / 0.77 \\
& M-NegCosSch(\textbf{ours}) &0.28 &0.18 / 0.38 &0.34 / 0.53 &0.28 &0.38 / 0.24 &0.39 / 0.34 \\
& P-NegCosSch(\textbf{ours}) &0.21 &0.13 / 0.31 &0.26 / 0.35 &1.08 &1.37 / 0.55 &2.07 / 1.18 \\\midrule
\multirow{3}{*}{SupCon + LS} &Constant &0.2 &0.24 / 0.32 &0.44 / 0.38 &0.69 &1.03 / 0.55 &1.35 / 0.85 \\
& M-NegCosSch(\textbf{ours}) &0.38 &0.36 / 0.17 &0.53 / 0.18 & 0.19 &0.41 / 0.37 &0.31 / 0.43 \\
& P-NegCosSch(\textbf{ours}) &0.32 &0.62 / 0.4 &0.78 / 0.57 &0.53 &1.77 / 0.42 &1.82 / 0.51 \\\midrule
\multirow{3}{*}{\parbox{1.5cm}{ARPL (\citet{chen2021adversarial})}}& Constant & 0.17 &0.55 \ 0.56 &0.37 \ 0.47 &0.46 &0.47 \ 0.56 &0.41 \ 0.58 \\
& M-NegCosSch(\textbf{ours}) &0.19 &0.85 / 0.42 &0.54 / 0.26 &0.42 &0.62 / 0.59 &0.67 / 0.51 \\
& P-NegCosSch(\textbf{ours}) &0.17 &0.87 / 0.45 &0.49 / 0.3 &0.47 &0.63 / 0.58 &0.62 / 0.57 \\\midrule
&&\multicolumn{3}{c}{\textbf{SCars}} &\multicolumn{3}{c}{\textbf{TinyImageNet}}  \\\midrule
\multirow{3}{*}{CE (w/o LS)}& Constant &0.1 &0.52 / 0.49 &0.51 / 0.52 &0.2 &0.21 &0.23 \\
&M-NegCosSch(\textbf{ours}) &0.09 &0.22 / 0.51 &0.26 / 0.55 &0.27 &0.27 &0.32  \\
& P-NegCosSch(\textbf{ours}) &0.1 &0.29 / 0.62 &0.21 / 0.59 &0.38 &0.32 &0.4 \\\midrule
\multirow{3}{*}{CE + LS} &Constant&0.1 &0.26 / 0.22 &0.25 / 0.25 &0.25 &0.24 &0.21  \\
& M-NegCosSch(\textbf{ours}) &0.08 &0.3 / 0.32 &0.33 / 0.29 &0.15 &0.22 &0.19  \\
& P-NegCosSch(\textbf{ours}) &0.14 &0.26 / 0.3 &0.34 / 0.31 &0.29 &0.37 &0.36 \\\midrule
\multirow{3}{*}{\parbox{1.5cm}{SupCon (w/o LS)}} &Constant&0.19 &0.12 / 0.45 &0.25 / 0.39 &0.18 &0.01 &0.17 \\
& M-NegCosSch(\textbf{ours}) &0.15 &0.18 / 0.18 &0.18 / 0.24 &0.2 &0.04 &0.15  \\
& P-NegCosSch(\textbf{ours}) &0.14 &0.2 / 0.29 &0.2 / 0.32 &0.37 &0.08 &0.33 \\\midrule
\multirow{3}{*}{SupCon + LS}& Constant &0.06 &0.18 / 0.36 &0.19 / 0.35 &0.25 &0.16 &0.27 \\
& M-NegCosSch(\textbf{ours}) &0.1 &0.2 / 0.38 &0.23 / 0.33 &0.16 &0.16 &0.09  \\
& P-NegCosSch(\textbf{ours}) &0.05 &0.19 / 0.25 &0.19 / 0.25 &0.15 &0.12 &0.14 \\\midrule
\multirow{3}{*}{ARPL}& Constant& 0.47 &0.08 / 0.92 &0.27 / 0.79 &0.1 &0.13 &0.12 \\
& M-NegCosSch(\textbf{ours}) & 0.34 &0.13 / 0.5 &0.07 / 0.37 &0.37 &0.29 &0.24 \\
& P-NegCosSch(\textbf{ours}) & 0.19 &0.12 / 0.67 &0.18 / 0.7 &0.17 &0.25 &0.17 \\
\bottomrule
\end{tabular}
\end{table}

\subsection{Performance on Vision Transformer}\label{sec:ViT}
We evaluate our proposed NegCosSch on the CUB benchmark using a pretrained tiny vision transformer (ViT) \citep{wu2022tinyvit}, with the results are presented in Table \ref{tab:ViT}. The improvements observed in the table, along with previous results from VGG and ResNet-based architectures, confirm the applicability of our proposed TSs across a range of model architectures.

\begin{table}[h]\centering
\caption{Performance on CUB with a Tiny ViT architecture.}\label{tab:ViT}
\scriptsize
\begin{tabular}{lrrrr}\toprule
method &Accuracy (\%) &AUROC (\%)& OSCR (\%) \\\midrule
Const. (baseline) &90.83 &91.41 / 79.88 &82.99 / 72.59 \\
M-NegCosSch (ours) &\textbf{91.07} &\textbf{92.06} / \textbf{80.67} &\textbf{83.79} / \textbf{73.51} \\
P-NegCosSch (ours) &\textbf{91.2} &\textbf{92.3} / \textbf{80.04} &\textbf{84.1} / \textbf{73.03} \\
\bottomrule
\end{tabular}
\end{table}

\subsection{Ablations on $(\tau^+,\tau^-)$}\label{sec:result1}

\begin{figure}[t]
  \centering
  \begin{subfigure}[b]{0.49\textwidth} 
    \centering
    \includegraphics[width=\textwidth]{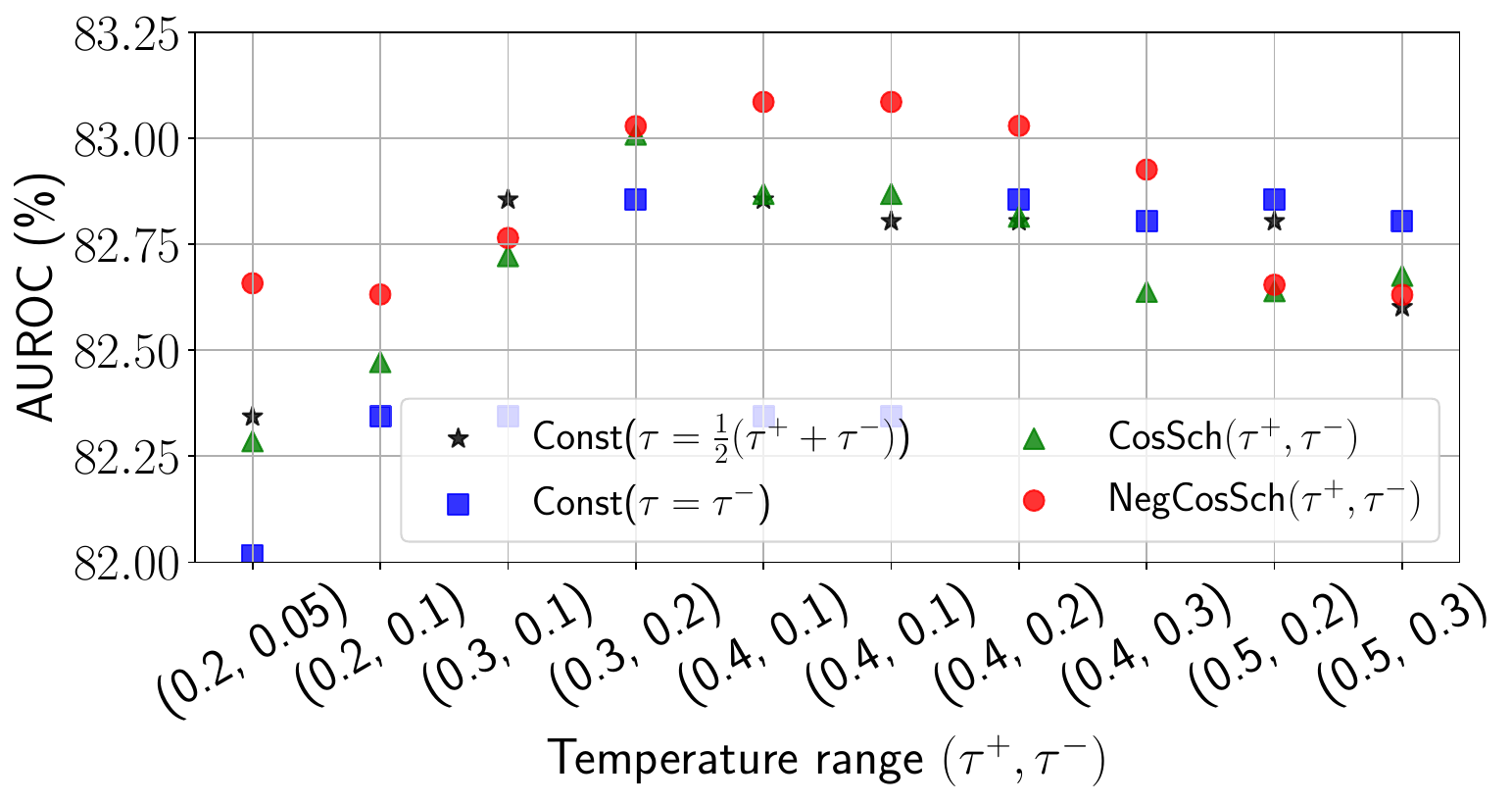}
    \caption{TinyImageNet} % 
    \label{fig1:subfig_d} 
  \end{subfigure}
  \begin{subfigure}[b]{0.49\textwidth} 
    \centering
    \includegraphics[width=\textwidth]{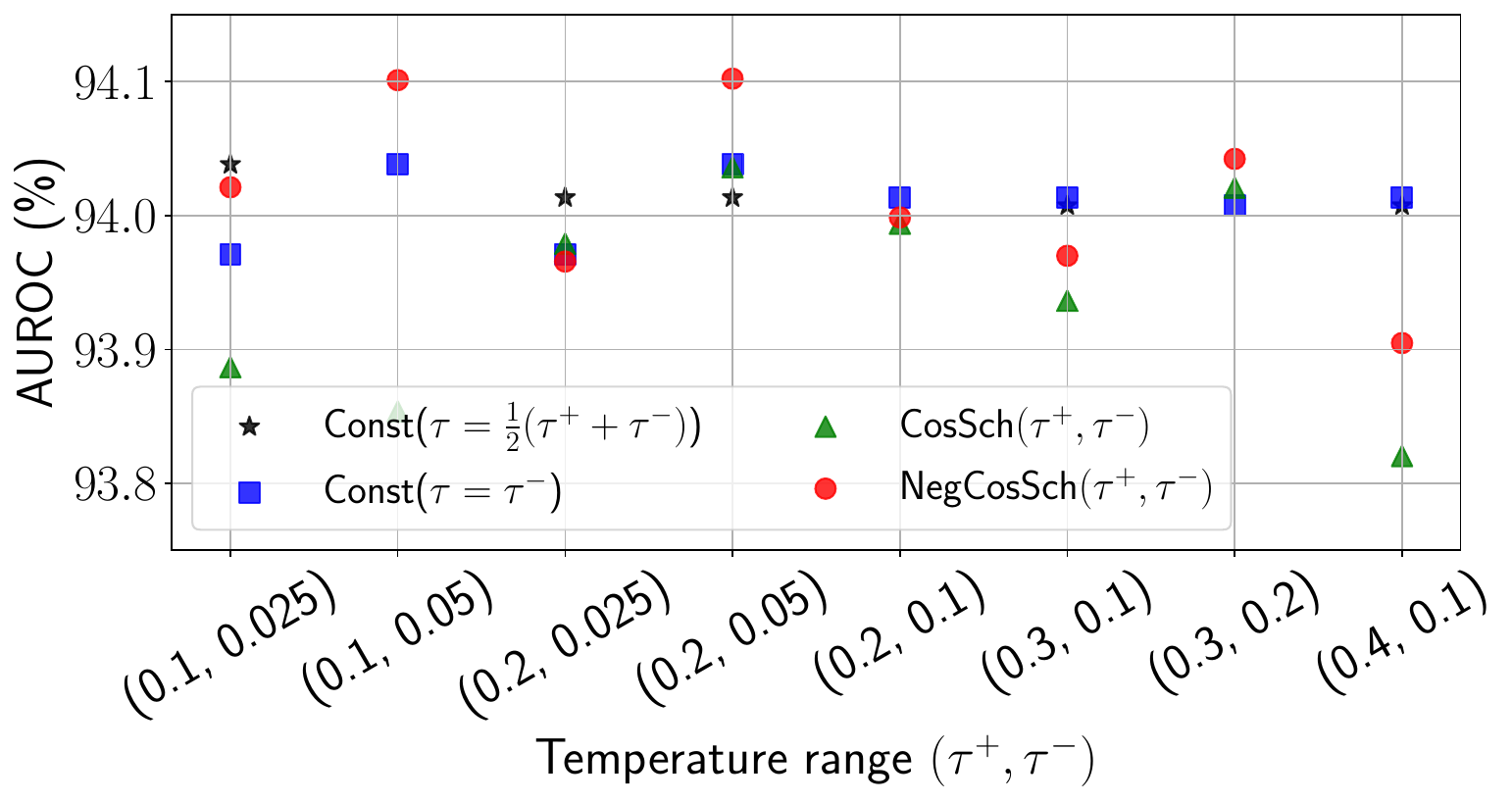}
    \caption{CIFAR10} % 
    \label{fig1:subfig_a} 
  \end{subfigure}\\
  \begin{subfigure}[b]{0.49\textwidth} 
    \centering
    \includegraphics[width=\textwidth]{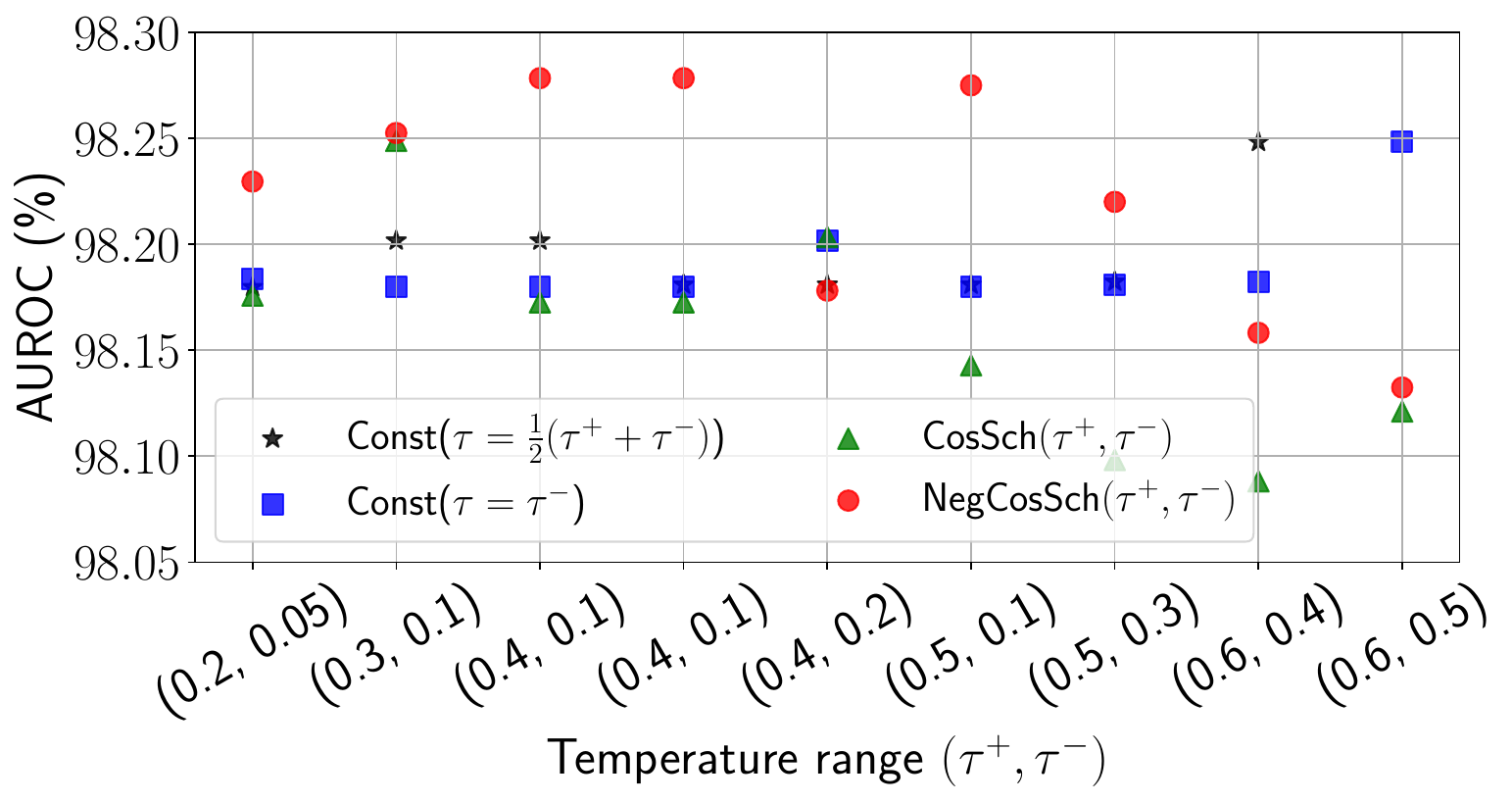}
    \caption{CIFAR+10} % 
    \label{fig1:subfig_b} 
  \end{subfigure}
  \begin{subfigure}[b]{0.49\textwidth} 
    \centering
    \includegraphics[width=\textwidth]{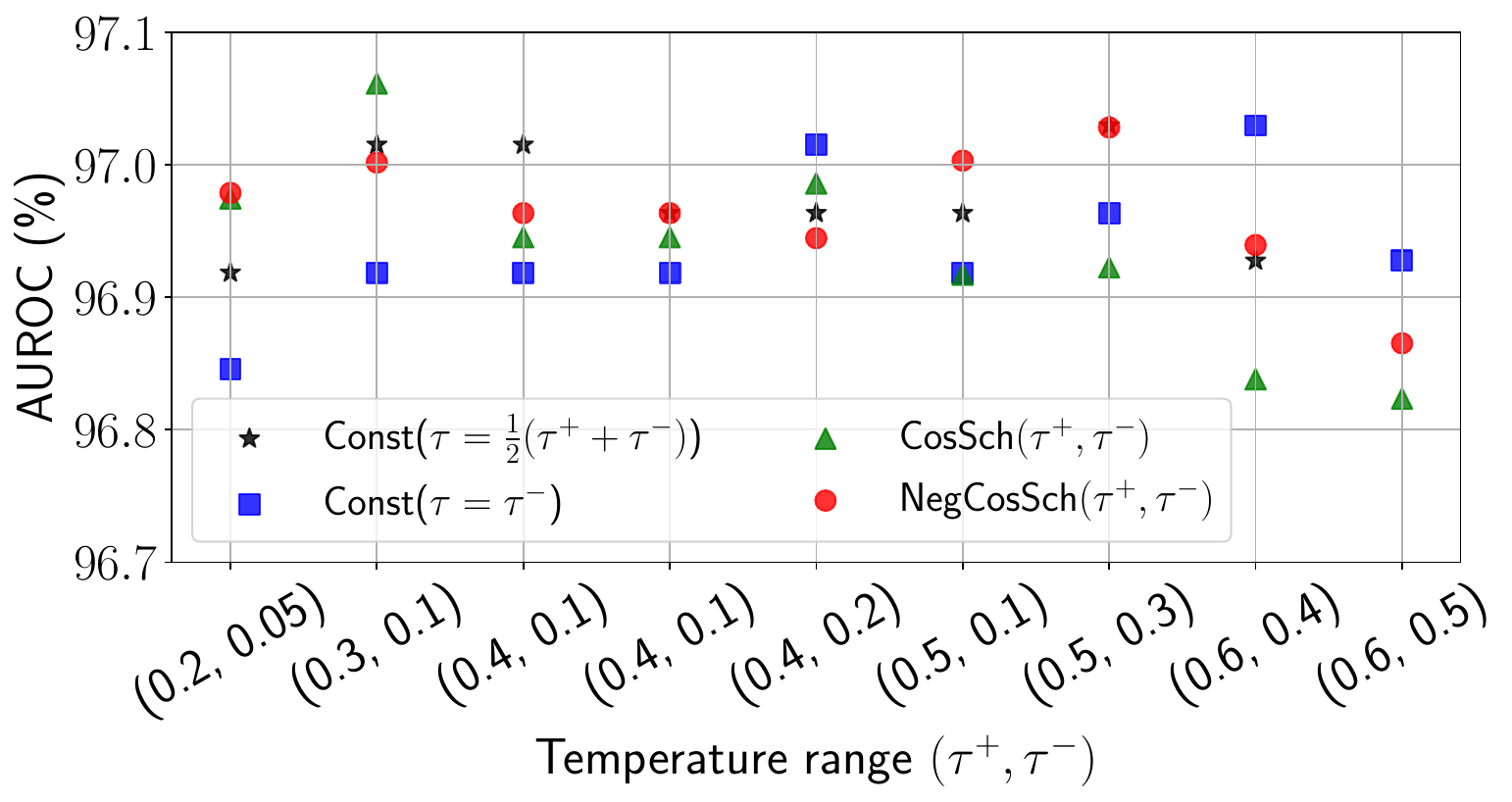}
    \caption{CIFAR+50} % 
    \label{fig1:subfig_c} 
  \end{subfigure}
  \caption{Open Set AUROC of different TSs for the SupCon loss on the regular OSR benchmarks.}\label{fig:1}
\end{figure}

Figure \ref{fig:1} presents the OSR performance of our NegCosSch along with the regular cosine TS and constant temperatures in SupCon loss on the regular benchmarks. We vary $(\tau^+,\tau^-)$ from $\mathbb{T}^2_{SupCon}$ and compare $\mathcal{T}_{\text{NegCosSch}}(\tau^{+},\tau^{-},P)$ with $\mathcal{T}_{\text{CosSch}}(\tau^{+},\tau^{-},P)$, $\mathcal{T}_{\text{Const}}(\tau= \text{nearest}(\frac{1}{2}[\tau^{+}+\tau^{-}]))$\footnote{By `nearest', we mean a nearest temperature is chosen from $\mathbb{T}_{\text{SupCon}}$. } and $\mathcal{T}_{\text{Const}}(\tau=\tau^-)$. 
The objective of the quadruplet-wise comparisons is to determine if our proposed TS outperforms a regular cosine TS, a constant temperature set to the midpoint of $(\tau^{+},\tau^{-})$, or set to $\tau^-$ with various pairs of $(\tau^+,\tau^-)$.
We observe that for the CIFAR10, CIFAR+10 and TinyImageNet benchmarks, our proposed TS yields a better open set AUROC than CosSch and constant temperatures for most of the quadruplet comparisons. We find the improvements or degradations to be insignificant for the CIFAR+50 benchmark with the highest AUROC found for $\mathcal{T}_{\text{CosSch}}(0.3,0.1)$. By observing the best performances of our NegCosSch, we formulate the strategy mention in Section \ref{Model_training} for choosing $(\tau^+,\tau^-)$.

\begin{comment}
\begin{table}[!h]\centering
\scriptsize
 \setlength{\tabcolsep}{4pt}
\caption{Closed set accuracy and open set AUROC (in \%) for the SupCon baseline without and including our proposed TS on the regular OSR benchmarks. The numbers in parenthesis represent the standard deviation across the 5 defined splits.
%Open set performance (AUROC \%) comparison with the other methods in the literature.
}\label{tab:app_comparison}
\begin{tabular}{lrrrrrrrr}\toprule
 &\multicolumn{2}{c}{CIFAR10} & \multicolumn{2}{c}{CIFAR+10} & \multicolumn{2}{c}{CIFAR+50} &  \multicolumn{2}{c}{TinyImageNet}\\
Methods &Acc. & AUROC &Acc. & AUROC &Acc. & AUROC &Acc. & AUROC \\\midrule

%CE (Baseline) (\citet{vaze2022openset}) & & 93.6 & &97.9 & &96.5 & &83 \\
%CE + NegCosSch (\textbf{ours}) & & \textbf{93.76}& & 97.88& & 96.32 & & \textbf{83.47} \\\midrule
Baseline & 96.95(1.47)& 94.04(0.97) & 98.05(0.60)& 98.25(0.27) & 98.13(0.53)&97.03(0.23) & 85.14(2.77) &82.86(1.78) \\
%1.47	0.97	0.60	0.27	0.53	0.23	2.77	1.78
NegCosSch (\textbf{ours})& 96.91(1.40)& \textbf{94.10}(0.85) & 98.02(0.54)& \textbf{98.28}(0.40) &98.10(0.50)& \textbf{97.03}(0.20) &\textbf{85.18}(3.42)& \textbf{83.09}(2.10) \\
%1.40	0.85	0.54	0.40	0.50	0.20	3.42	2.10
\bottomrule
\end{tabular}
\end{table}
\end{comment}

\subsection{Ablations on $P$ and $k$ in $\mathcal{T}_{\text{GCosSch}}$}
We perform an ablation study on $P$ in $\mathcal{T}_{\text{NegCosSch}}$ using TinyImageNet with the SupCon loss for different pairs of $(\tau^+,\tau^-)$ and the open set AUROC are presented in Table \ref{tab:diff_P}. We observe that different choices of $P$ produce similar OSR performance.

We also compare the open set AUROC on TinyImageNet among different values of $k$ in $\mathcal{T}_{\text{GCosSch}}$. The temperatures are set to: $\tau^+=0.4$ and $\tau^-=0.1$. From Table \ref{tab:diff_sche}, we observe that the open set AUROC increases with the value of $k$, with the highest AUROC observed for $k=1$ or our NegCosSch.

\begin{table}[!h]
\begin{minipage}[h]{0.48\textwidth}
\centering
\scriptsize
\caption{Open set AUROC (\%) on TinyImageNet for different values of $P$ in $\mathcal{T}_{\text{NegCosSch}}$.
}\label{tab:diff_P}
\begin{tabular}{lrr}\toprule
%\begin{tabular}{lr}\toprule
$(\tau^+,\tau^-)~~\backslash~~~~P\rightarrow$& 100 & 200\\\midrule
(0.3,0.2) &83.10 &83.03 \\
(0.4,0.1) &82.91 &83.09 \\
(0.4,0.2) &82.99 &83.03 \\
\bottomrule
\end{tabular}
\end{minipage}\hfill
\begin{minipage}[h]{0.48\textwidth}
\centering
\scriptsize
\caption{Open set AUROC on TinyImageNet for different values of $k$ in $\mathcal{T}_{\text{GCosSch}}$.
}\label{tab:diff_sche}
\begin{tabular}{lr}\toprule
%\begin{tabular}{lr}\toprule
Schedule & AUROC (\%)\\\midrule
%Const & 82.86\\
%random & 83.05\\ 
%Step-up & 82.93\\
%Half-NegCosSch & 82.98\\
$\mathcal{T}_{\text{GCosSch}}(k=0)$ or $\mathcal{T}_{\text{CosSch}}$ & 82.87\\
$\mathcal{T}_{\text{GCosSch}}(k=0.25)$ & 82.93\\
$\mathcal{T}_{\text{GCosSch}}(k=0.50)$ & 82.99\\
$\mathcal{T}_{\text{GCosSch}}(k=0.75)$ & 83.03\\
$\mathcal{T}_{\text{GCosSch}}(k=1)$ or $\mathcal{T}_{\text{NegCosSch}}$ & \textbf{83.09}\\
\bottomrule
\end{tabular}
\end{minipage}
\end{table}

\subsection{Performance on Prototypical Contrastive Learning}
We also show results with our NegCosSch on TinyImageNet using the prototypical contrastive (ProtoCon) learning in Table \ref{tab:protocon}. The ProtoCon loss is recently used by \citet{bahavan2025sphor, li2025unlocking} for OSR. We observe improvements for all three metrics.

In ProtoCon, instead of contrasting an anchor representation with another sample, we contrast with the prototypes of known classes. We randomly initialize one prototype per known class $\{p_c\}_{c=1}^C$. The loss function forces all representations of the same class to lie near its prototype and to move away from other prototypes, which is given as: 
\begin{equation}
L_{ProtoCon} = - \frac{1}{|I|}\sum_{i \in I}  \text{log} \frac{\text{exp}(\text{sim}(l_i,  p_{\tilde{y}_i}) /\tau)}{\sum_{c =1}^C \text{exp}(\text{sim}(l_i,  p_c) /\tau)} 
\end{equation}
We update the prototypes on the fly at each iteration $t$. $\sigma$ is the learning rate for prototypes.
\begin{eqnarray}
p_{c}^t = 
\begin{cases}
p_{c}^{t-1} & \text{; if } |\{i \in I: \tilde{y}_{i} = c\}| = 0 \\
(1-\sigma) p_{c}^{t-1} + \frac{\sigma}{|\{i \in I: \tilde{y}_{i} = c\}|} \sum_{\{i \in I: \tilde{y}_{i} = c\}} l_i,   &\text{; otherwise} \\
\end{cases}
\end{eqnarray}

\begin{table}[!htp]\centering
\scriptsize
\caption{Performance on TinyImageNet using prototypical contrastive learning}\label{tab:protocon}
%\resizebox{ extwidth}{!}{ % use this if the table is too large
\begin{tabular}{lrrrr}\toprule
TS &Accuracy (\%) &AUROC (\%) &OSCR (\%) \\\midrule
Const. (baseline) &85.36 &82.79 &70.53 \\
NegCosSch &\textbf{85.72} &\textbf{83.04} &\textbf{71.12} \\ %M-NegCosSch
%P-NegCosSch &84.52 &\textbf{82.82 }&69.94 \\
\bottomrule
\end{tabular}
\end{table}

\subsection{Performance on the CIFAR Benchmarks}
Here, we evaluate our periodic NegCosSch using the SupCon loss on the CIFAR benchmarks -- such as CIFAR10, CIFAR+10 and CIFAR+50 and the results are presented in Table \ref{tab:app_comparison}. The values of $\tau$ for $\mathcal{T}_{\text{Const}}$ are chosen as $0.05,0.5$, and $0.4$ respectively with hyperparameter tuning for the CIFAR10, CIFAR+10, and CIFAR+50 benchmarks and the values of $(\tau^+,\tau^-)$ in our TS are $(0.2,0.05), (0.4,0.1)$, and $(0.5,0.3)$ respectively. We observe that the closed set accuracy is similar to the baseline methods when we include our TS on these benchmarks, whereas we gain slight improvements in the open set performance. The open set performance depends on the nature of the unknown classes and their semantic similarity with the known classes. We suspect that the benefits of our TS reduce when the number of training classes is relatively small, which occur in the CIFAR benchmarks.
%. We find that the improvements over the baselines decrease, with a slight drop in closed set accuracy, when there is a small number of training classes, which occur in the CIFAR10, CIFAR+10, and CIFAR+50 benchmarks.}
For example, there are only 6 training classes in CIFAR10 and 4 training classes in the CIFAR+10 and CIFAR+50 benchmarks. Moreover, the OSR AUROC on the CIFAR+10 and CIFAR+50 benchmarks are $>97\%$ with tuned constant temperature baselines, leaving only a little scope for improvements.
However, as mentioned before, we observe significant improvements both for the open set and closed set performance on the TinyImageNet and the SSBs, where they have a larger number of training classes.

\begin{table}[!h]\centering
\scriptsize
 \setlength{\tabcolsep}{4pt}
\caption{Closed set accuracy, open set AUROC and OSCR (in \%) for the SupCon baseline without and including the proposed NegCosSch on the CIFAR benchmarks. %The numbers in parenthesis represent the standard deviation across the 5 defined splits.
%Open set performance (AUROC \%) comparison with the other methods in the literature.
}\label{tab:app_comparison}
\begin{tabular}{lrrrrrrrrr}\toprule
 &\multicolumn{3}{c}{CIFAR10} & \multicolumn{3}{c}{CIFAR+10} & \multicolumn{3}{c}{CIFAR+50} \\
Methods &Accuracy & AUROC &OSCR&Accuracy & AUROC&OSCR &Accuracy & AUROC &OSCR \\\midrule

%CE (Baseline) (\citet{vaze2022openset}) & & 93.6 & &97.9 & &96.5 & &83 \\
%CE + NegCosSch (\textbf{ours}) & & \textbf{93.76}& & 97.88& & 96.32 & & \textbf{83.47} \\\midrule
Const. (Baseline) & 96.95& 94.04 &91.13& 98.05& 98.25 &96.32& 98.13&97.03&95.21 \\
%1.47	0.97	0.60	0.27	0.53	0.23	2.77	1.78
NegCosSch (\textbf{ours})& 96.91& \textbf{94.10} &\textbf{91.18}& 98.02& \textbf{98.28} &\textbf{96.33}&98.10& 97.03&95.17 \\
%1.40	0.85	0.54	0.40	0.50	0.20	3.42	2.10
\bottomrule
\end{tabular}
\end{table}

\section{Related Works (Continued)} 
%Here, we provide a comprehensive list of OSR methods from the literature with their OSR performance on the regular benchmarks in Table \ref{tab:methods}. The results are taken either from \citet{vaze2022openset} or from the individual work.
Here, we discuss the recent OSR methods.
\citet{wang2024exploring} propose to extract diverse features from multiple experts with an attention diversity regularization to ensure the attention maps are mutually different. \citet{zhou2024contrastive} propose a framework with contrastive training for classification and implement an additional VAE for reconstruction to compute an unknown score based on intermediate features. 
\citet{yang2024dynamic} propose a self-learning framework for test time adaptation. 

Another line of work utilizes data augmentation. For example, \citet{jia2024revealing} propose an asymmetric distillation to feed the teacher model with extra data through augmentation, filtering out the wrong prediction from the teacher model and assigning a revised label to them to train the student model. The method in \citet{wang2025backmix} augments the dataset by mixing the foreground of images with different backgrounds. \citet{xuinformed} propose new data augmentation with the help of visual explanation techniques, such as the LayerGAM to mask out the activated areas so that models can learn beyond the discriminative features.

The other methods are based on contrastive learning with different regularization. For example, \citet{xu2023contrastive,li2024all} train models with contrastive loss, sample mix up and label smoothing for better representation learning. 
\citet{bahavan2025sphor} also propose a prototypical contrastive loss to pull all samples to its class prototype and push away the prototypes of other classes. \citet{li2025unlocking} propose a regularization inspired from the neural collapse perspective -- the closed set classes are aligned with a simplex equiangular tight frame geometric structure. 
Recent works by \citet{zhou2024decoop, hua2025openworldauc} introduce open world prompt tuning methods that improve a vision language model's performance in an open-world scenario to make better predictions from a mix of known and unknown classes.

Although the recent methods aim for better representation learning, some of them achieve this through feeding more data to the model with augmentation. On the other side, a few recent OSR methods do not use the same experiments settings maintained in most of works in the literature. For example,
\citet{wang2025backmix,wang2024exploring,jia2024revealing} use different backbone models for evaluation, which makes it harder to compare their methods with others.

\section{Implementation}\label{implementation}
Each model is trained on a single NVIDIA-RTX2080Ti GPU requiring from 2 to 32 hours depending on the model and the dataset. 
Our implementation utilizes Python (v3.7) and PyTorch (v1.12), accelerated with CUDA (v11.3) and cuDNN (v8.2). Our codes are mostly built on top of the code-base by \cite{vaze2022openset} and the implementation of SupCon loss is taken from the official GitHub page by \citet{khosla2020supervised}. Our periodic NegCosSch schedule can be integrated into any existing loss with a few lines of codes as the following:

\begin{verbatim}
import math
class GCosineTemperatureScheduler:
  def __init__(self, t_p=2.0,t_m=0.5, P=200,shift=1.0,epochs=600):
    self.t_p = t_p
    self.t_m = t_m
    self.epochs = epochs
    self.P = P
    self.s = shift
    self.e = int(self.epochs - 0.5 * self.s * self.P)
  def get_temperature(self, epoch):
    if(t<self.e):
      t = self.t_m + (self.t_p - self.t_m) * 
        (1+ math.cos(2*math.pi* (epoch-self.s * self.P/2)/self.P))/2 
    else:
      t = self.t_p
    return t

if(args.temperature_scheduling):
  TS=GCosineTemperatureScheduler()
for epoch in range(1,N_epochs+1):
  if(args.temperature_scheduling):
    criterion.temperature = TS.get_temperature(epoch)
  # rest of the code
  ... ...
\end{verbatim}

\end{document}